# Development of a magnetorheological hand exoskeleton featuring a high force-to-power ratio for enhanced grip endurance

**Wenbo Li[1,2], Xianlong Mai[1,2], Ying Li[1,2], Weihua Li[3], Shiwu Zhang[1,2] Lei Deng[4*], Shuaishuai Sun [1,2*]**

[1]CAS Key Laboratory of Mechanical Behavior and Design of Materials, Department of Precision Machinery and Precision Instrumentation, University of Science and Technology of China, Hefei, Anhui 230026, China.
[2]Institute of Humanoid Robots, School of Engineering Science, University of Science and Technology of China, Hefei, Anhui 230026, China.
[3]School of Mechanical, Materials, Mechatronic and Biomedical Engineering, University of Wollongong, New South Wales 2522, Australia.
[4]School of Electrical, Computer and Telecommunications Engineering, University of Wollongong, Wollongong, New South Wales, 2522, Australia.

Email: sssun@ustc.edu.cn; leideng@uow.edu.au

∗ Authors to whom any correspondence should be addressed.

## Abstract

Hand exoskeletons have significant potential in labor-intensive fields by mitigating hand grip fatigue, enhancing hand strength, and preventing injuries. However, most of the traditional hand exoskeletons are driven by motors, whose output force is limited in the constrained installation conditions. Besides, they also come with the disadvantages of high power consumption, complex and bulky assistive systems, and high instability. In this work, we develop a novel hand exoskeleton integrated with innovative magnetorheological (MR) clutches that offers a high force-to-power ratio to improve grip endurance. The clutch features an enhanced structure design, a micro roller enhancing structure, which can significantly boost output forces. The experimental data demonstrate that, when it is supplied with 2V, the clutch can deliver a peak holding force of 381.15 N—55 times that when no voltage is provided (7 N). In this scenario, it only consumes 1.38 W, yielding a force-to-power ratio of 256.75N/W, which is 2.35 times higher than the best-reported actuator used for hand exoskeletons. This capability enables the designed MRHE to provide approximately 419.79 N support force for gripping. The designed MR hand exoskeleton is highly integrated, comprising an exoskeleton frame, MR clutches, a control unit, and a battery. Evaluations through static grip endurance tests and dynamic carrying and lifting tests confirm that the MR hand exoskeleton can effectively reduce muscle fatigue, extend grip endurance, and minimize injuries. These findings highlight its strong potential for practical applications in repetitive tasks such as carrying and lifting in industrial settings.

## 1. Introduction

AS the most active and interactive part of the upper extremity, the human hand plays a crucial role in both daily life and work. Most activities of daily living (ADLs) [1] and many occupation-related activities involve numerous scenarios where the hand is required to perform static gripping or repetitive and forceful grasping movements to lift and hold heavy objects [2,3]. However, prolonged and high-intensity gripping can lead to muscle fatigue, and long-term physical labor under poor ergonomic conditions predisposes individuals to work-related musculoskeletal disorders (WRMDs) [4], especially in labor-intensive fields such as construction, logistics, and manufacturing. Hence, it is important to improve hand grip endurance, where workers are often required to perform gripping tasks under physically demanding conditions. Although traditional hand protection and support devices, such as wrist supports, gloves, and grip pads, can offer some relief, they generally do not substantially reduce the strain on the hands during prolonged use because of their poor load capacity and compatibility. In this context, the emergence of hand exoskeleton technology offers a promising solution due to its excellent versatility and ergonomic friendliness [5]. By integrating appropriate support mechanisms within a hand exoskeleton, it is possible to enhance grip endurance, reduce muscle strain, and improve work efficiency, all while minimizing the risk of hand injuries associated with prolonged labor.

Over the past two decades, hand exoskeletons have been extensively studied for rehabilitation, assistance with daily activities, and task-specific training [6–14]. Existing designs of hand exoskeletons generally use four types of actuators [15]: electric motor-based [16,17], pneumatic [18], and shape memory alloy-based (SMA-based) actuators [19]. Electric motors, including both linear [20–22] and rotational [23–26] motors, are widely available and easy to control. However, they need transmission systems such as gearboxes to amplify actuating force or torque, which increases space consumption. Besides, the limited space in hand-exoskeletons restricts the improvement of power density. Pneumatic exoskeletons offer higher power-to-weight ratios and enhanced mechanical compliance [27,28], but they require external air compression systems, which are often bulky, noisy, and have limited operational durations [29]. While offering a high power-to-volume ratio and compact design, SMA-based exoskeletons can only deliver small forces with low response speeds [30] and relatively low energy efficiency [31]. The rope-driven passive exoskeleton in [32] attempts to alleviate finger fatigue, but its effect on grip endurance is minimal. Each actuator type has its own advantages and limitations; however, all face the challenge of providing sufficient support force for high-load applications. These actuators are typically positioned directly or through specific transmission mechanisms on the exoskeleton frame of the hand. Due to the limited space available in the hand exoskeleton, the actuator size is constrained, which restricts its output force and limits the capability of the exoskeleton to deliver enhanced support. Most reported hand exoskeletons are primarily designed to assist with daily activities and rehabilitation. They typically exhibit a relatively low force-to-power ratio and provide limited support force. Therefore, they may be inadequate for tasks demanding significant grip strength and prolonged gripping, such as lifting heavy objects. For instance, the elastic actuators of the hand exoskeleton in [33] can only generate an assistive force of 20 N per finger, while the linear actuators in [34] provide a maximum of around 10 N per finger, limiting their application in the prolonged gripping task with large grip strength. Therefore, it is crucial to design a mechanism and hand exoskeleton capable of providing sufficient support force while conserving energy, especially for applications involving carrying large items. Achieving this goal requires the development of novel structures or materials with a high force-to-power ratio to serve as the core components of the exoskeleton.

Magnetorheological (MR) materials have recently shown significant potential in advancing semi-active exoskeletons with a high force-to-power ratio [35–40]. Among these materials, magnetorheological grease (MRG) has attracted particular interest due to its superior anti-settling properties and low fluidity. MRG consists of micron-sized ferrous particles suspended in a non-magnetic grease matrix. When an external magnetic field is activated, the rheological properties of MRG undergo rapid and reversible changes. At the microscopic level, the ferrous particles within the carrier fluid form magnetic chains aligned with the magnetic field lines. MR devices that use MRG as a force transmission medium, such as MR clutches or MR dampers [41–44], have been developed and demonstrated to be effective in transmitting high forces. During the transmission process, MRG experiences shearing or squeezing, which results in a dynamic 'break-recover' of the magnetic chains at the microscopic scale, enabling stable and continuous torque transmission. The transformation of MRGs rheological state is characterized by low power consumption, ease of control, reversibility, and rapid response time in the millisecond range [45,46]. These merits make it especially suitable for robotic applications that demand high load capacity and low power consumption.

To meet the urgent demand for hand exoskeletons with powerful grip support, this study proposes a highly integrated magnetorheological hand exoskeleton (MRHE) system designed to enhance grip strength and endurance while relieving hand fatigue. The MRHE is fully wearable, portable, autonomously controlled, and characterized by a high force-power ratio. It highly integrated an exoskeleton frame, MRG clutches (MRGCs), a control board, and a battery into a compact system. Four linear MRG clutches (MRGCs) were employed to support fingers, excluding the thumb, as they are primarily responsible for executing gripping tasks. The MRGC was designed with a performance enhancement structure consisting of a miniature roller assembly and MRG. Adding the micro rollers into the traditional MR gap as a performance enhancement structure, the wedge effect forming around the rollers substantially enhances the maximum holding force of the clutch under low power consumption, achieving a high force-to-power ratio with a compact structure. The MRGC is able to provide a peak holding force of 368.24 N at a power consumption of 1.38 W, yielding a force-to-power ratio of 276.18 N/W within its recommended usage range of input voltage. These endow the MRHE with the capability to deliver over 400 N of additional support to the human hand with just 5.68W, considering the four MRGCs and other devices integrated into this system of power consumed (Entire system), thereby reducing muscle fatigue during prolonged, high-intensity gripping. Due to its portability, high force-to-power ratio, and high level of integration, the MRHE is particularly well-suited for hand exoskeleton applications in physical tasks that require substantial load capacity.

The main contributions of this study are highlighted as follows:

1) Designed a novel linear MR grease clutch incorporating an innovative micro-roller contact pair structure that induces a squeeze-strengthen effect in MRG, substantially enhancing output force compared to traditional linear MR clutches and achieving a superior force-to-power ratio over conventional actuators.
2) Developed a highly integrated hand exoskeleton conveying mechanical architecture, autonomous control systems, and powering units, enabling efficient and reliable grip assistance.
3) Experimentally validated the MRHE in industrial scenarios, demonstrating its superior effectiveness in enhancing grip endurance and reducing wearer fatigue during high-load tasks, such as weight-carrying and weight-lifting.

The subsequent sections are structured as follows: Section 2 presents the overall design of the hand-exoskeleton prototype, including the exoskeleton frame, MRGC design, and electronic control systems and control strategy of MRHE. Section 3 evaluates the proposed MRHE by analyzing the internal magnetic field, electrical parameters, and mechanical characteristics of its MRGC, followed by an exploration of the relationship between the holding force of the MRGC and the grip support force of the MRHE. Section 4 outlines the static grip experiments and applications, wherein a human wears the MRHE to verify the improvement in grip endurance. Lastly, section 6 provides a discussion with final remarks.

## 2. Design of the hand exoskeleton prototype

As shown in Figure 1(a), the proposed hand exoskeleton prototype consists of an exoskeleton frame, four MRGCs, a control board, and a battery. The MRGCs are connected to the exoskeleton frame via hinges to provide support force for the MRHE. The control ports of the four MRGCs are linked to the microcontroller board via wires. Powered by the battery, the control board autonomously regulates the holding force of the MRGCs according to the control algorithm. The highly integrated MRHE is secured to the human body using Velcro straps, as shown in Figure 1(b).

### *2.1 Hand exoskeleton frame*

As shown in Figure 2, the MRHE system consists of an exoskeleton base, four linear MRGCs, finger mechanisms, and two pressure sensors, with the four MRGCs and finger mechanisms attached to the base using hinges. These components work together to enhance the exoskeleton's load-bearing capacity, thereby improving the user's grip ability and endurance.

As illustrated in Figure 2(b), the finger mechanism consists of the linkage-hinge and fingerstall structures. The fingerstall structure comprises three sequentially connected segments, linked by two hinges that align with the finger joints to enable natural and comfortable movement. Each segment is made of a metal cover and a Velcro strip. The metal covers are fabricated using aluminum powder sintering, a metal 3D printing technique that ensures precise control over the material's properties. This process creates a lightweight yet robust shell that offers essential support while maintaining flexibility and an appropriate fit. The Velcro straps secure fingers to the exoskeleton, allowing easy adjustments to accommodate various hand sizes and quick attachments or removal for practical daily use.

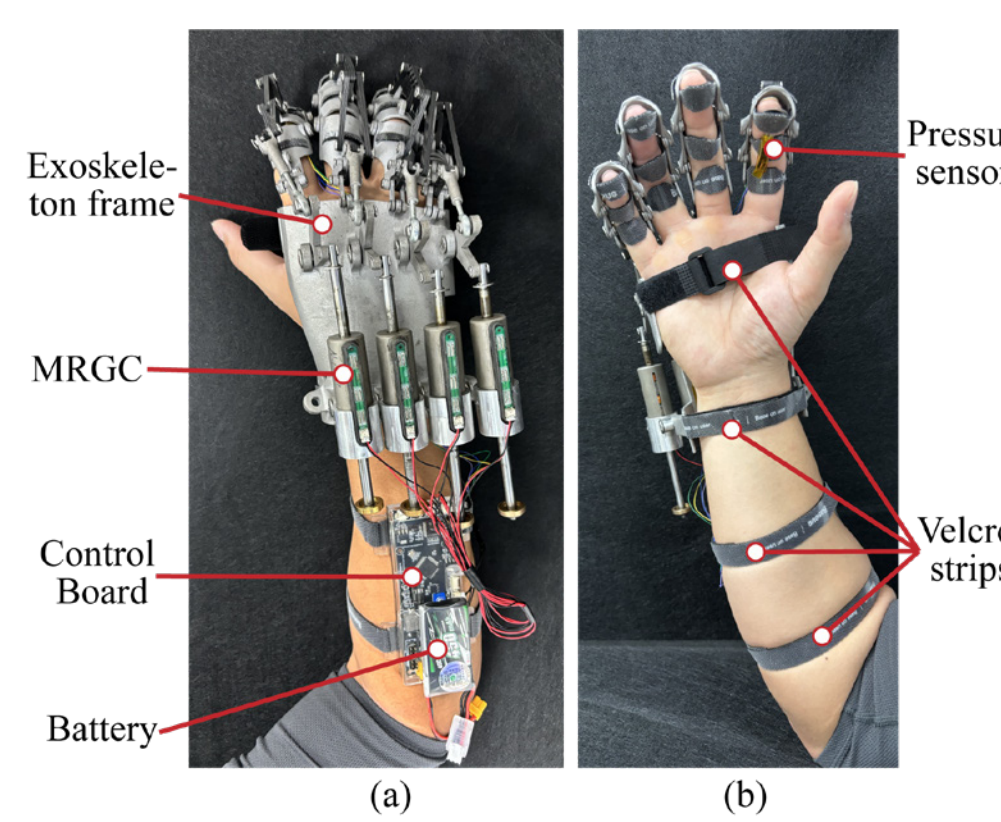

**Figure 1.** Demonstration and composition of the hand exoskeleton system. (a) Dorsal-side view of the system. (b) Palm-side view of the hand exoskeleton system.

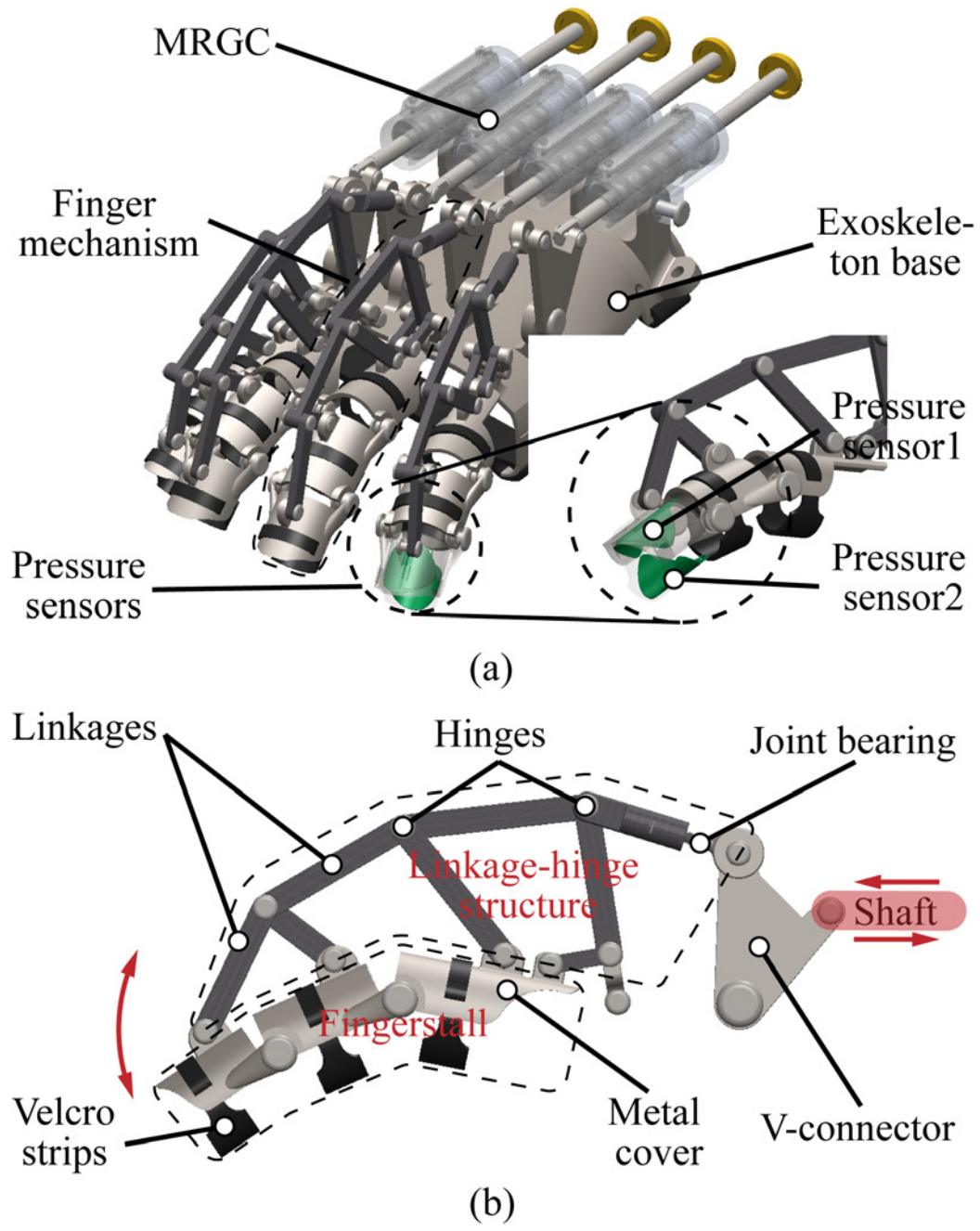

**Figure 2.** Structural Design and Key Components of the Hand Exoskeleton. (a) Overall structure of the hand exoskeleton. (b) Details of the finger structure.

The linkage-hinge structure comprises linkages, joint bearings, and hinges. On the one side, it connects fingerstall segments, while on the other, it links to the shaft of the MRGC via a V-connector. This connection converts finger movements into the linear motion of the MRGC shaft, enabling the MRGCs to provide the support force for the finger mechanism.

Additionally, the design incorporates two pressure sensors (RP-C18.3-ST, LEGACT, China) located at the fingertip of the index finger. The sensors are responsible for providing real-time pressure feedback to detect the user's grip intention, enabling intuitive operation and seamless interaction between the user and the exoskeleton.

### *2.2 MRGC prototype*

#### *2.2.1 Performance enhancement structure*

The supporting force that the MRHE can provide is directly dependent on the holding force generated by the MRGC. As shown in Figure 3(a), conventional MR clutch configures MRG or MRF in gaps between two parallel plates or cylinders, where damping forces are generated by the shear stress of the MRG or MRF, and the magnitude is controllable via regulating the magnetic field strength. However, the damping force generated by this method is very limited [37,41,47], especially in scenarios when the design size is constrained. To address this issue, we proposed a structure named Micro Roller Contact Pair (MRCP) as shown in Figure 3(b). Compared to conventional structures, additional rollers and retainers are placed in the enlarged gap alongside MRG between the static and motion walls. When the motion wall moves, the roller rotates while the retainer remains still. This structure is inspired by the wheel chock used to prevent vehicle wheels from rotating. As shown in Figure 3, when the MRG performs high viscosity under a magnetic field, MRG accumulates in the triangular region, defined as the hardened zone, forming wedge shapes similar to a wheel chock. This wedge effect generates significant resistance, effectively restraining the roller's rotation and producing a large damping force to oppose the motion of the walls.

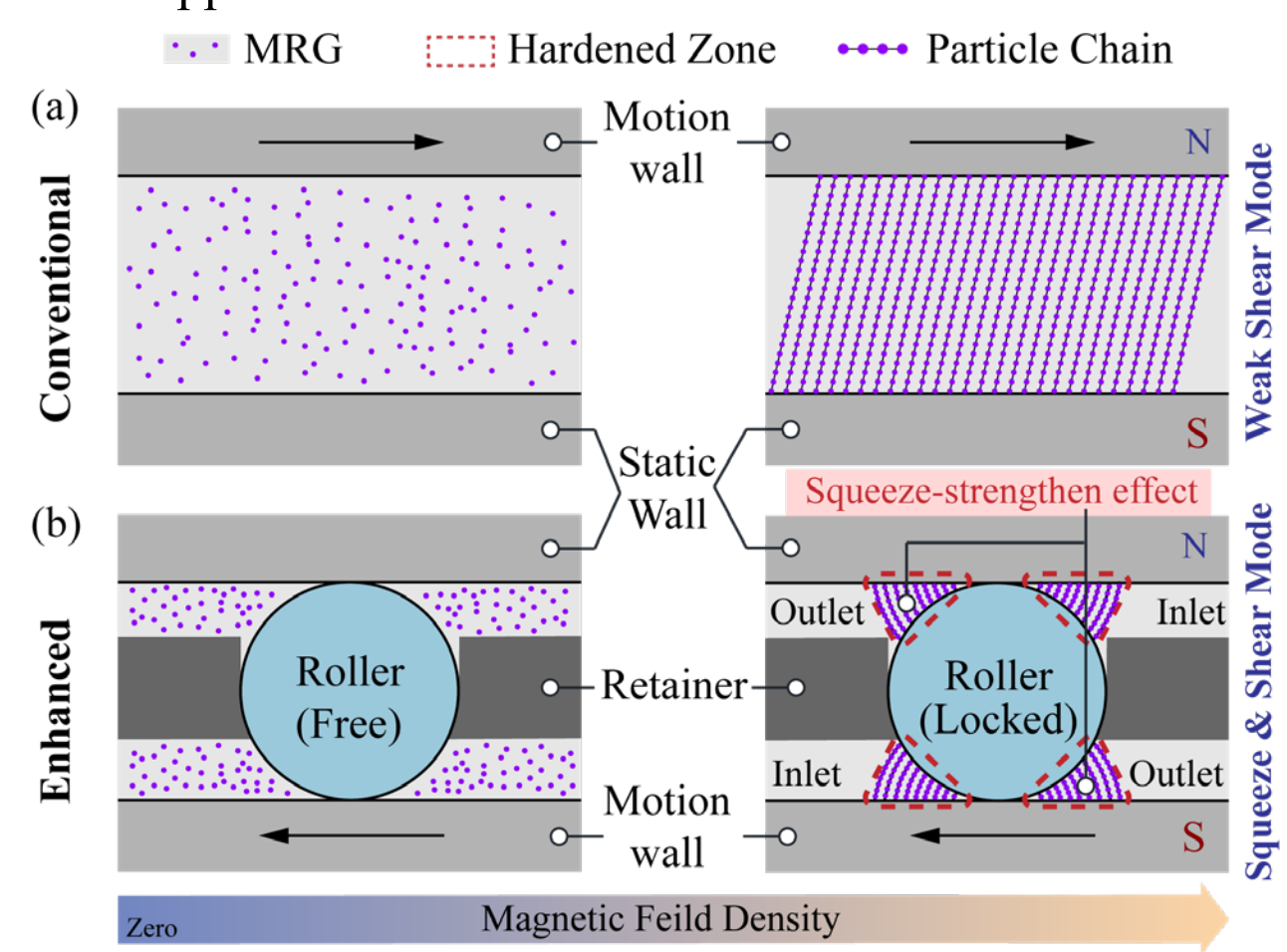

**Figure 3.** Performance enhancement structure-MRCP. (a) Conventional structure. (b) Enhanced structure.

The detailed work principle of the MRCP is demonstrated in Figure 4. When no magnetic field is applied, the ferrous particles in the MRG disperse randomly in the carrier (Figure 4(a)), and the device exhibits the characteristic of low resistance. In this case, the MR clutch is in the disengagement state. As shown in Figure 4(b), in the engagement state, when a magnetic field is applied, the ferrous particles form magnetic chains, resulting in the hardening of the MRG. If an external force is applied to the shaft, the hardened parts would work in both squeeze and shear mode, generating the resistance force, as shown in Fig. 4(c). Furthermore, according to previous studies [48–51], under compression, the shear stress of MRG or MRF is capable of generating a significant shear force and ten times the yield stress compared to a pure shear mode. With this squeeze-strengthen effect, the generated force from MRCP can be enhanced. When the external force meets the maximum resistance force this structure can generate, the MRG will transfer from the elastic region to the yielding state (Figure 4(d)), generating relative motion among the MRCP. After the roller motion, the MRG in the hardened zone, under the guidance of the local magnetic field and the entrainment effect [52] of the grease, will continuously converge to the inlet side to maintain the resistance.

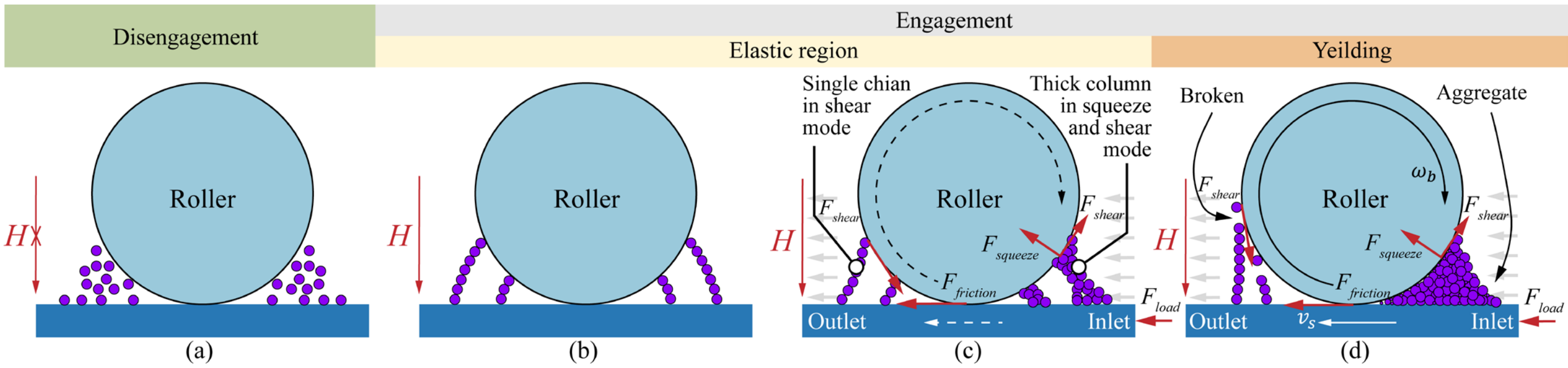

**Figure 4.** Resistance generation process. (a) Initial. (b) Magnetized. (c) Motion trend. (d) In motion.

Besides the sheer force, the magnetic chains at the outlet are squeezed, generating a squeeze force, as shown in Figure 4(c). These hardened zones should be deconstructed to make roller motion possible, resulting in the resistance force. However, as shown in Figure 4(d), after the MRG is deconstructed by the roller motion, the magnetic field can easily reform the magnetic chains and harden the MRG again. Apparently, since the cross-section of the roller is circular, most of the resistance arises from the deformation process of the MRG during the roller's motion. This resistance force can be constructed by the theory of plasticity. Under the action of the applied magnetic field, the yield stress of the MRG can be controlled by adjusting its stiffness. The deformation process of breaking the MR solid phase requires the application of an external force on the moving part, where the resistance force is proportional to the applied magnetic field. The remainder of the resistance force is the friction force, which depends on the manufacturing tolerances of the components, operating conditions, and the properties of the MRG. Thus, the total resistance generated by the MRCP structure comprises three parts: squeeze force, shear force, and friction force, while the squeeze force contributes most due to the wedge effect.

With the designed MRCP structure, the holding force of the MRGC increases significantly under the same magnetic field strength, demonstrating the characteristic of a high force-to-power ratio. The braking torque generated during the extrusion-deformation process of the MRG is the primary contributor to this feature. Furthermore, the desired resistance can be achieved within a smaller space by increasing the number of MRCPs, significantly reducing the overall volume of the device.

### *2.2.2 Mechanical design of the MRGC*

Based on the analysis of the previous section, we proposed a miniature linear MRGC for MRHE by utilizing the MRCP structure, as shown in Figure 5(a). The MRGC comprises one shaft, two roller retainers, one outer sleeve, two end caps, two magnetically conductive rings, a printed circuit board (PCB) connector, three coils, and three coil carriers.

The MRGC is embedded in the exoskeleton frame. The end of the shaft is connected to the end of the finger transmission mechanism of the frame, and the remaining part of MRGC is connected to the base of the frame through a hinge. This configuration ensures that the force acting on the exoskeleton frame is ultimately transferred to the MRGC. Thus, the holding force generated by the MRGC directly impacts the supporting force provided by the MRHE.

To maximize the holding force within the limited space of the exoskeleton frame, we optimized the volume of the MRGC and increased the number of MRCPs as much as possible. Each MRGC contains 80 MRCPs, divided into two groups. The rollers in each group are fixed in a retainer made of aluminum alloy, as shown in Figure 5(a). The shaft (motion wall) and magnetic-conducting ring (static wall) sandwich the rollers, forming the MRCP structure. The material of soft electromagnetic iron (Type: DT4C), which possesses high magnetic permeability and low coercivity, is selected for the outer sleeve and magnetic-conducting ring to reduce energy loss and improve the dynamic response of the device. To ensure sufficient mechanical strength for the payload, the shaft and the rollers are made of AISI 1010 steel. As shown in Figure 5(b), MRG with a mass fraction of 92.4% fills the remaining voids of the MRGC. The rheological characteristics of the MRG, measured using a rheometer (MCR502, Anton Paar Corp.), are presented in Figure 6. The relationships between shear stress and magnetic field density, and between viscosity and magnetic field density can be described by the following two polynomial fitting equations.

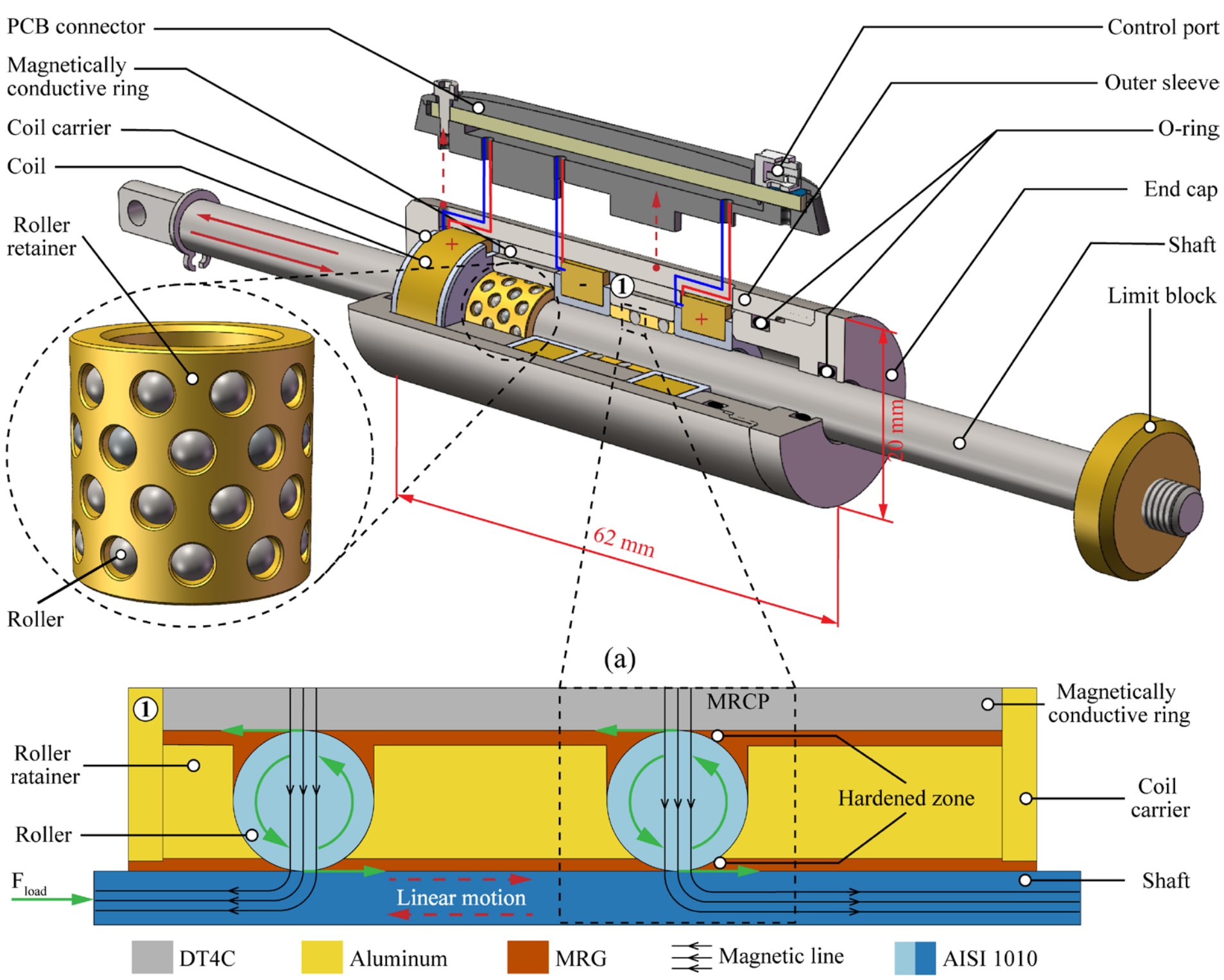

**Figure 5.** Structural composition of the MRGC with miniature rollers. (a) Overall structural schematic. (b) Enlarged view of roller structure section.

$$\tau_y(B) = 0.7992 - 6.635B + 224.9B^2 - 114.3B^3 \quad (1)$$

$$\eta(B) = 0.08252 - 0.7294B + 22.71B^2 - 11.51B^3 \quad (2)$$

To generate and control the magnetic field at the MRCP, three sets of coils (copper wire, 2 mm) are used. The coils are arranged alternately with the magnetic conductive ring, ensuring the magnetic field covers all MRCPs and generates magnetic field lines perpendicular to the surface of the wall. The polarities of adjacent coils are opposites, ensuring that the magnetic poles activated between two adjacent coils have the same polarity. Under the control of varying magnetic field density, magnetic chains form between the surfaces of the roller and the walls, creating MRG hardened zones of varying intensities. The outermost layer of the MRGC uses an outer sleeve and an end cap made of DT4C to constrain and guide the magnetic field. At both ends of the shaft movement, O-rings are used to seal the MRG to prevent leakage. Three coils are connected in parallel through a PCB port, providing a control interface for controlling the coils. When the magnetic field is activated by the coil, the harden zone shown in Figure 5(b) will prevent the relative motion among the MRCP. The external force will transmit from the shaft to magnetically conductive ring via the MRG and rollers. This working mechanism allows the MRGC to achieve high force output in a compact structure. The relevant geometric and mass parameters can be found in Table 1.

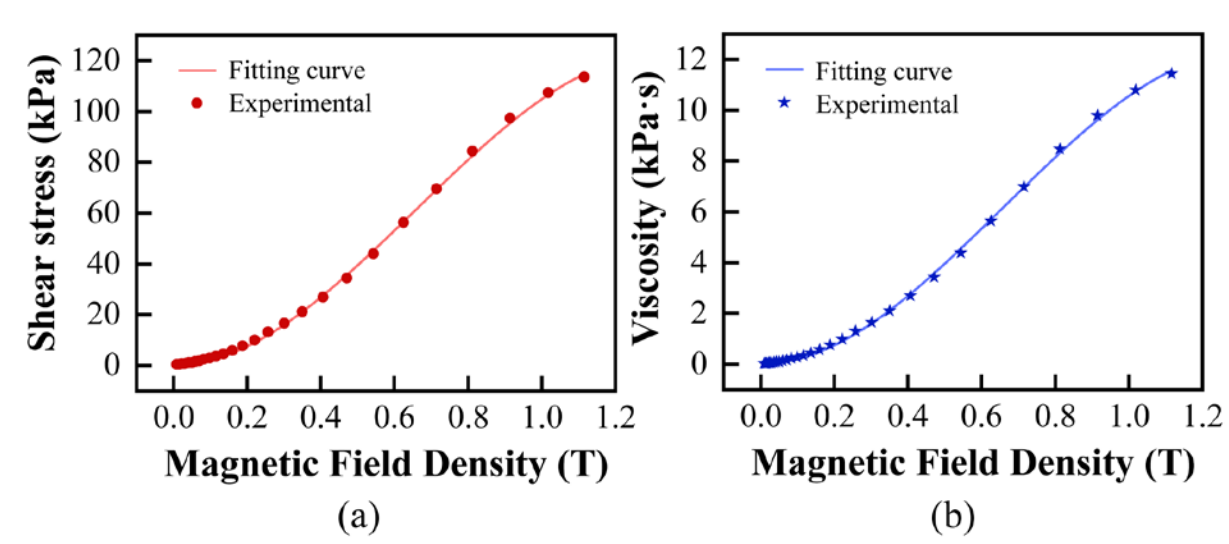

**Figure 6.** Rheological characteristics of MRG. (a) Shear stress-magnetic field density fitting curve. (b) Viscosity-magnetic field density fitting curve.

**Table 1.** Mechanical parameters of MRGCs

| Parameters | Value | Unit |
|---|---|---|
| Weight of MRGC | 153.79 | g |
| Length of MRGC $l_m$ | 62 | mm |
| Diameter of MRGC | 20 | mm |
| Shaft stroke length | 68 | mm |
| Shaft diameter $d_s$ | 6 | mm |
| Shaft length | 150 | mm |
| Roller retainer length $l_r$ | 9 | mm |
| Roller diameter | 1.5 | mm |
| MR gap $g$ | 0.175 | mm |

### *2.2.3 Electrical parameters*

The electrical parameters of the coils are an important reference when designing MRGC control circuits and strategies. We primarily measured the internal resistance $R_c$, the series equivalent inductance $L_s$, and the quality factor $Q$ of the MRGC. The results are presented in Table 2. The resistance was measured using a digital multimeter (86B, VICTOR, China), while $L_s$ and $Q$ were measured with a digital bridge (4092C, VICTOR, China) at a frequency of 100 kHz.

**Table 2** Electrical parameters of MRGCs

| MRGC index | $L_s(\mu H)$ | $Q$ | $R_c(\Omega)$ |
|---|---|---|---|
| 1 | 147.522 | 3.511 | 2.91 |
| 2 | 146.728 | 3.536 | 2.87 |
| 3 | 146.663 | 3.592 | 2.88 |
| 4 | 148.769 | 3.470 | 2.92 |

The average resistance of the four MRGCs used in the MRHE is 2.90 $\Omega$. The average $L_s$ is 147.42 $\mu H$, and the average $Q$ is 3.53. The relatively high $Q$ value indicates that the MRGC exhibits good control performance at a control frequency of 100 kHz. Subsequently, the control voltage of the MRGC is regulated using a Pulse Width Modulation (PWM) signal with a frequency of 100 kHz.

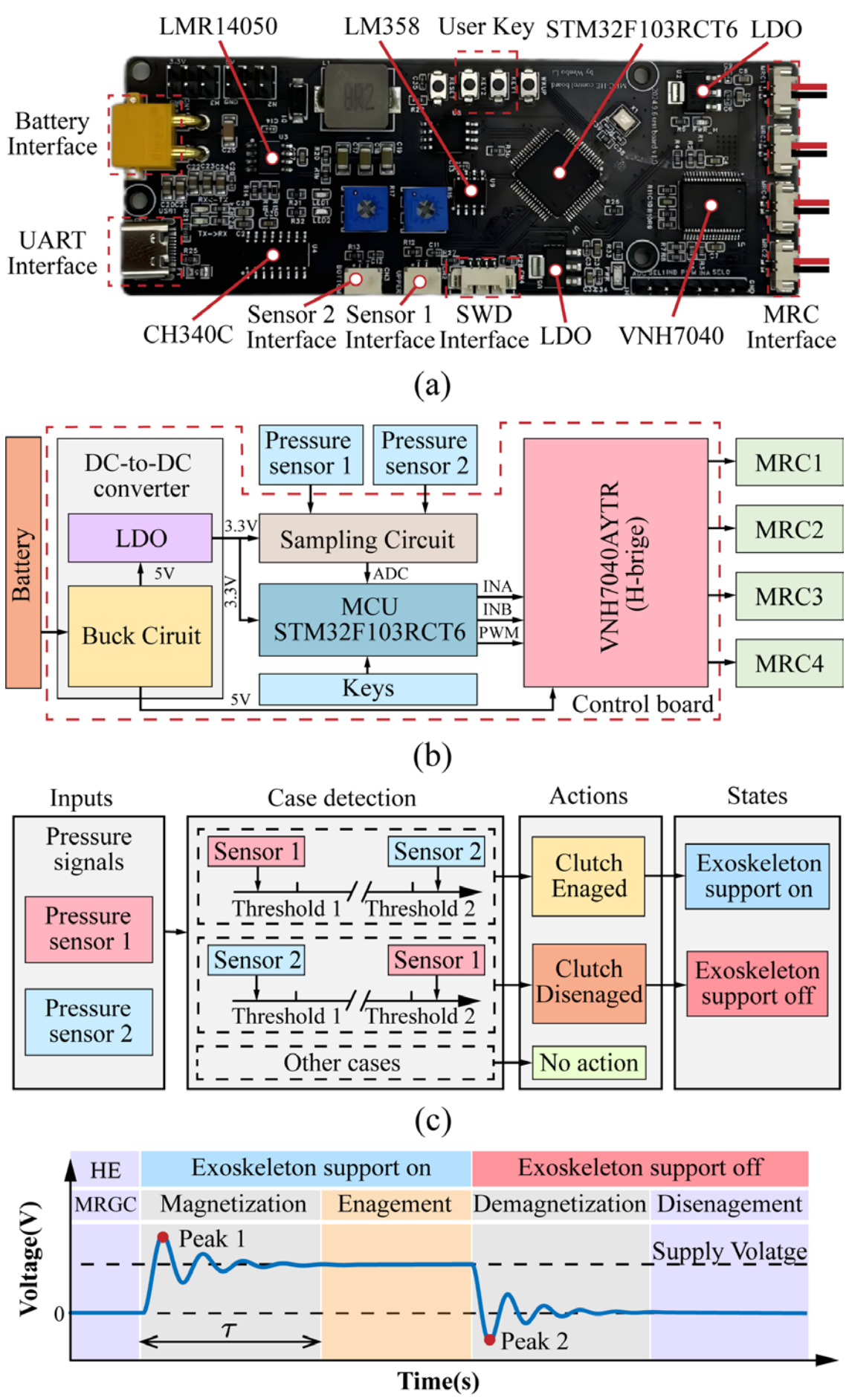

**Figure 7.** Control Circuit Board and Logic of the Hand Exoskeleton System. (a) Structural composition of the control board. (b) Hardware circuit topology diagram. (c) Control logic flowchart. (d) Voltage command to MRGC and corresponding operational states.

### *2.3 Electrical design and control*

To enable seamless interaction between the wearer and the exoskeleton, a highly integrated control board, as shown in Figure 7(a), was developed to control the MRGCs using real-time pressure feedback from the index finger. This board merges data reception, analysis, and control functions into a single unit, enhancing the wearability and compact design of the hand exoskeleton.

The control board consists of a 32-bit microcontroller unit (MCU), STM32 (STM32F103RCT6), an H-bridge driver circuit (VNH7040) to control the operation of four MRGCs (connected in parallel), a sampling circuit with the single supply dual operational amplifier (LM358) to measure pressure sensor signal, a universal Asynchronous Receiver/Transmitter (UART) communication circuit (CH340C) to transmit debug message, a DC-to-DC converter with a buck circuit (LMR14050) and linear regulator (LDO), one set of keys to set different supporting force, a Serial Wire Debug (SWD) interface to download the program, and corresponding interfaces to connect the board with a supply battery, two pressure sensor, and four MRGCs.

The working principle of this control board is illustrated in Figure 7(b). Supplied by a Li-Po battery (GS4502S45, 7.4V, 450mAh, ACG, 32g, China), the buck circuit powers the H-bridge driver circuit with a 5 V voltage while the sampling circuit and the MCU are powered by a voltage of 3.3 V with the help of the LDO. During the operation, signals from the pressure sensors are processed through the sampling circuit, which converts the resistance changes of the sensor into analog signals to be analyzed by the MCU. Then, commands for the H-bridge to control the MRGCs are made by the MCU based on the designed control strategy.

The hand exoskeleton is designed to provide support force while the wearer is gripping an object to relieve fatigue. Upon releasing the object, the exoskeleton should turn off to ensure the hand's motion remains unrestricted. To achieve the control target, a control logic, presented in Figure **7**(c), is implemented to identify the wearer's intention, which is based on the feedback from the two pressure sensors: one located on the finger dorsum (sensor 1) and the other on the finger pad (sensor 2). Those pressure values are compared with two thresholds, which are determined and adjusted based on actual wear and experience. In this system, we set threshold 2 at 2.0 V as the higher one and threshold 1 at 1.0 V as the lower one. When the value from sensor 2 exceeds threshold 2 and sensor 1's value is below threshold 1, the control logic identifies that the wearer is gripping an object. In response, the MRGCs are engaged, and the exoskeleton is activated to provide the necessary support force. Conversely, when the intention to release the object is detected-indicated by sensor 2's value reducing below threshold 1 and sensor 1's value rising above threshold 2, the MRCGs are disengaged, which disables the exoskeleton and allows the wearer to release the object without any impedance. In other scenarios, the exoskeleton remains off, ensuring the hand can move unrestrictedly. The control logic is implemented in C and downloaded to the control board via the SWD interface, which runs the control program independently.

The ability of MRGCs to switch quickly and smoothly between engaged and disengaged states is crucial for the MRHE to provide instant support when gripping an object and eliminate undesired impediments when releasing it. In this regard, a powering method, as illustrated in Figure 7(d), is applied to achieve quick magnetization and demagnetization. During the transition of these two states, sinusoidal signals with reduced amplitude, as expressed by the following equation, are used to control the supplied voltages of MRGCs:

$$V_s(t) = V_{target} + \left(V_{target} - V_{current}\right)(exp\left(-\frac{mt}{\tau}\right) \\ sin\left(2\pi\frac{mt}{\tau} + pi/2\right)) \quad (3)$$

where $V_s$ denotes the voltage during the state-switching process, $V_{target}$ is the target voltage for the power supply after the switching, $V_{current}$ denotes the voltage before the switching, $\tau$ is the required time, and $m$ denotes the number of sine periods.

## 3. Valuation of the hand exoskeleton prototype

### *3.1 Analysis of the MRG clutch*

In this section, we first validated the effectiveness of the design in enhancing and controlling the magnetic field within the MRCP through magnetic field simulation. Then, the mechanical characteristics of the MRGC were evaluated and analyzed using a Mechanical Testing System (MTS). Finally, a simple kinetic model was built to calculate the theoretical supporting force that the MRHE can provide.

#### *3.1.1 Magnetic simulation*

To validate the magnetic flux density (MFD) and effectiveness of the magnetic field lines in hardened zones, simulations were performed using COMSOL Multiphysics to analyze the MFD distribution in the MRGC when it is energized, and the result is presented in Figure8.

Figure 8(a) and Figure 8(b) demonstrate the vector magnetograms and contour of the magnetic flux density of the MRGC under a coil current of 0.1667A. The coils are arranged with opposing polarities, working cooperatively to control and enhance the MFD around MRCP. As shown in Figure 8(a), magnetic flux circuits with opposing directions are induced by the adjacent coils, leading to enhanced magnetic flux density in the MRCPs between them. Reflected by the color, Figure 8(b) confirms that most of the magnetic flux is concentrated in the rollers due to their high magnetic flux permeability, whereas the roller retainers, being non-magnetic, show significantly low concentration. Detailed MFD in the hardened zone is provided in Figure 8(c). Magnetic induction lines in this zone facilitate the MRG in forming the magnetic lines connected between two surfaces, restricting the rotation of the rollers, and increasing the damping. Figure 8(d) shows the MFD of an indexed track arranged in Figure 8(c), whose beginning is tangent to the wall, and the end is located in the middle of the gap. It is evident that MFD increases as the current increases or as the location approaches the tangential point. Figure 8(e) provides the relationship between the MFD and the applied current of two coils, one on the shaft and the other on the outer sleeve (as indicated by Probs 1 and 2 in Figure 8(b)), and the average value of the indexed track. It is observed that the average MFD of the indexed track undergoes substantial changes within the range of 0-0.35 A, which serves as the test range in the latter mechanical test. A distinct inflection point is observed at 0.2 A in the shaft's MFD curve (Prob 1), indicating the onset of magnetic saturation. Moreover, all regions (shaft, out sleeve, and hardened zone) exhibit a near-linear relationship between MDF and current from 0 to 0.2 A, suggesting a corresponding linear holding force

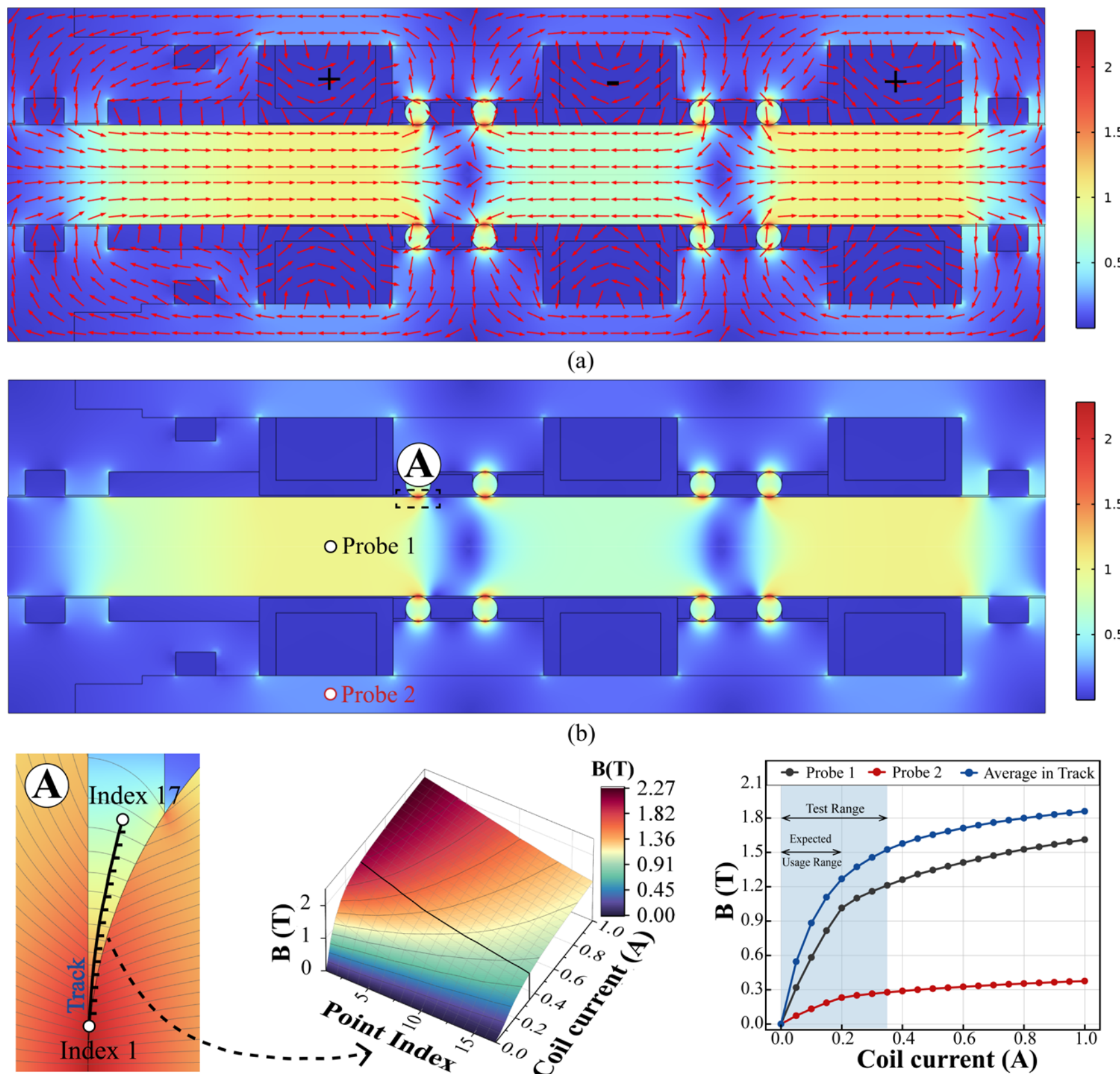

**Figure 8.** Simulation and Analysis of the Magnetic Field in the MRGC (a) Vector magnetograms at 0.1667A. (b) Contour of the magnetic flux density at 0.1667A. (c) The hardened zone around the roller with the track used to detect the MFD. (d) MFD along the track in different applied currents. (e) MFD at two probes and the average in the track under various input currents.

response for the MRGC, which is identified as the expected usage range of the MRGC.

### 3.1.2 Mechanical characterization

The peak holding force of the MRGC is a critical parameter, as it determines the clutch's maximum load-bearing capability. Therefore, the prototyped MRGC was characterized using a Material Testing System (MTS) (DN-W50KN, DAINA, China) to establish the correlation between the holding force and input voltage.

As illustrated in Figure 9(a) and (b), an MRGC is clamped to an MTS by two fixtures, which are made from stainless steel to avoid magnetic interference with the MRGC. The lower fixture secures the outer sleeve of the clutch, while the upper fixture is connected to the shaft. Under the excitation of the MTS, the shaft of the MRGC is forced to move cyclically between -10 mm and 0 mm at a speed of 10 mm/s. Meanwhile, the force and displacement data are recorded by the force sensor and the built-in displacement sensor. Suggested by the previous magnetic field analyzes, the applied current range is set as 0 - 0.35 A. Given that each MRGC consists of three parallel-connected coils with an average resistance of 2.9 Ω, the total current of the MRGC ranges from 0 A to 1.05A. Thus, in each test, voltages ranging from 0 to 3.0 V with an increment of 0.5 V are supplied to the MRGC by the DC Power.

The force-displacement response generated in a clockwise sequence is shown in Figure 9(c). It is observed that the response exhibited pronounced hysteresis loops, with their height increasing as the voltage rises. When the shaft changes direction, either the upper-left or the down-right corner, the force experiences a rapid initial increase, followed by a gradual decrease before the next direction change. This phenomenon indicates that a larger force is required in the pre-yield state of MRF than in the post-yield state. The data of peak force at various voltages is depicted in Figure 9(d). The force increases gradually at low voltages, showing a dramatic rise between 0 and 2.0 V according to the expected usage range in the magnetic simulation results. In this voltage range, the peak force increases 55 times from 7 N to 381.15 N. Beyond 2.0 V, the increase slows due to magnetic saturation. To describe this relationship, a fifth-order polynomial fitting was implemented to express the peak force $F_{peak}$ as a function of the supply voltage $V$. The resulting equation, equation (4), demonstrates a strong correlation between the experimental data and the polynomial curve, as shown in Figure 9(d).

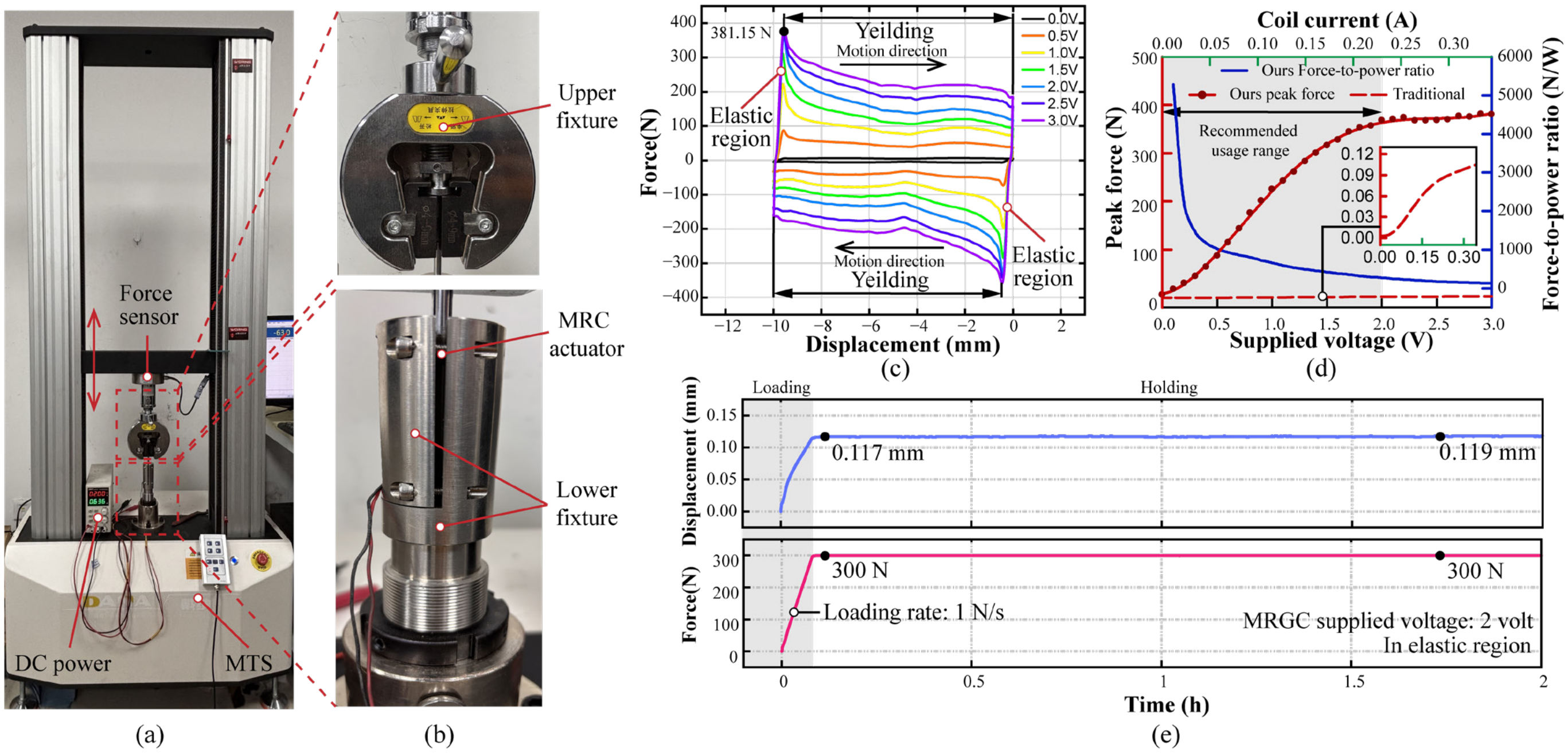

**Figure 9.** Experimental Setup and Results for the MRGC Testing. (a) Test rig. (b) Installation details. (c) Force-displacement relationship under different voltage conditions. (d) Correlation between the input voltage and the peak force (red line) of MRGC, as well as the force-to-power ratio (blue line). (e) Load stability test result of MRGC under 2 V supply and 300 N static load.

$$F_{peak}(v) = 8.336 + 35.412v + 375.650v^2 - 253.826v^3 + 59.948v^4 - 4.588v^5 \tag{4}$$

Figure 9(d) also presents the peak force—voltage curve (red dashed line) of a traditional linear MR clutch with the same dimensions as the proposed MRGC but without the unique MRCP structure. Detailed calculations are provided in the Supplementary Materials. It is seen that the traditional MR clutch can only achieve a peak force of 0.086 N at 2 V, whereas the MRGC with the MRCP structure reaches 381.15 N, which is 4432 times higher. This substantial improvement clearly highlights the effectiveness of the MRCP structure in enhancing the force output of the MR clutch.

In contrast to the peak force, the force-to-power ratio in Figure 9(d) demonstrates an approximate inverse proportionality trend with the input voltage. Specifically, it exhibits a rapid decline within the 0–0.5 V range. At 2 V, the system achieves a force-to-power ratio of 276.18 N/W, while at 3 V, this ratio decreases to 127.05 N/W. Considering the influence magnitude of the input voltage on holding force and the force-to-power ratio comprehensively, the recommended usage range of input voltage is determined as 0-2.0V.

Additionally, to assess the shear stability of the MRGC under prolonged load, we applied a static load of 300 N while supplying the MRGC with 2 V. The displacement initially stabilized at approximately 0.117 mm. As shown in Figure 9(e), during the two-hour static load test, the displacement increased slightly, from 0.117 mm to 0.119 mm. These results confirm that the MRGC maintains excellent shear stability under extended loading conditions, with minimal displacement variation, indicating its reliability for long-duration use in high-load applications of exoskeletons.

Figure 10 compares the performance of the proposed MRGC with reported counterparts used for hand exoskeletons [13,19,53–60]. In terms of holding force, the MRGC with 2.0V input voltage exceeds the best-performing actuator reported in [57] by a factor of 1.59 times. Furthermore, the MRGC demonstrated the highest force-to-power ratio, reaching 276.18 N/W, a 2.52-fold improvement over the actuator in [57]. The ability of the proposed MRGC to generate high holding forces with minimal energy input underscores its superior functionality and efficiency. This validates the effectiveness of the MRCP design in enhancing the MRGC's performance. With outstanding performance, the MRGC reported in this work holds significant potential for hand exoskeleton applications, particularly considering the spatial constraints associated with their installation. Since the actuating device largely determines the overall performance of exoskeletons, the proven superiority of MRGCs well justifies the merits of the proposed MRGC-based exoskeleton over traditional exoskeletons based on devices such as motors.

### *3.2 Kinetic model and analysis of the MRHE*

Based on the established relationship between the maximum holding force of the MRGC and the supply voltage, this section aims to obtain the support force of the MRHE by analyzing its topological structure. Due to the complexity of the kinematics calculations of the multi-link mechanism of the MRHE, a simplified analysis is adopted to assess the exoskeleton's performance. Specifically, we have chosen to analyze the hand's position when lifting heavy objects without the exoskeleton, focusing on the kinetics of a single finger for further simplification. Figure 11 establishes the kinetic model of a single finger of the MRHE. The kinetic model is simplified as the rigid body model. $l_i$ denotes the length of the abstract linkage between the two rotating shafts, with $i$ representing the member number. To further simplify, we assume that the external force acts on the component located at $l_4$.

To establish the relationships among all the forces (marked in red) including $F$, $T_1$, $T_3$, $T_4$, $T_5$, $T_7$, $T_8$, $T_9$, $T_{10}$ and $F_{MRC}$, four sets of right-handed Cartesian coordinates (marked in blue), labeled as $x_1o_1y_1$, $x_2o_2y_2$, $x_3o_3y_3$ and $x_4o_4y_4$, are established, with axes $y_2$, $y_3$ and $y_4$ aligning with $T_1$, $T_5$ and $T_8$, respectively. The origins of the four coordinates are fixed at the hinges which join members number 3 and 4, members number 1, 3, and 5, members number 5, 7, and 8, and members number 8, 9, and 10. The axis forces of the members are marked as $T_i$ and $T'_i$ The index number $i$ stands for its corresponding member number. $T_i$ and $T'_i$ are equal in magnitude but opposite in direction. The angular relationships are also marked in violet in Figure 11, where $\theta_1$, $\theta_3$, $\theta_3$, $\theta_7$, $\theta_8$, $\theta_9$, $\theta_{10}$, $\alpha$, $\beta$ are the angles between $T_3$ and $x_1$, $T_3$ and $x_2$, $T_5$ and $x_2$, $T_7$ and $x_3$, $T_8$ and $x_3$, $T_9$ and $x_4$, $T_{10}$ and $x_4$, $T_9$ and member number 13, $F_{MRC}$ and member number 14 respectively. In the coordinate $x_1o_1y_1$, the governing equations of the loading force for one finger, $F_{load}$, can be obtained according to equilibrium of forces:

$$F_{load} = T_3 \times sin\theta_1 \tag{5}$$

$$T_3 \times cos\theta_1 = T_4 \tag{6}$$

From equation (3), we have

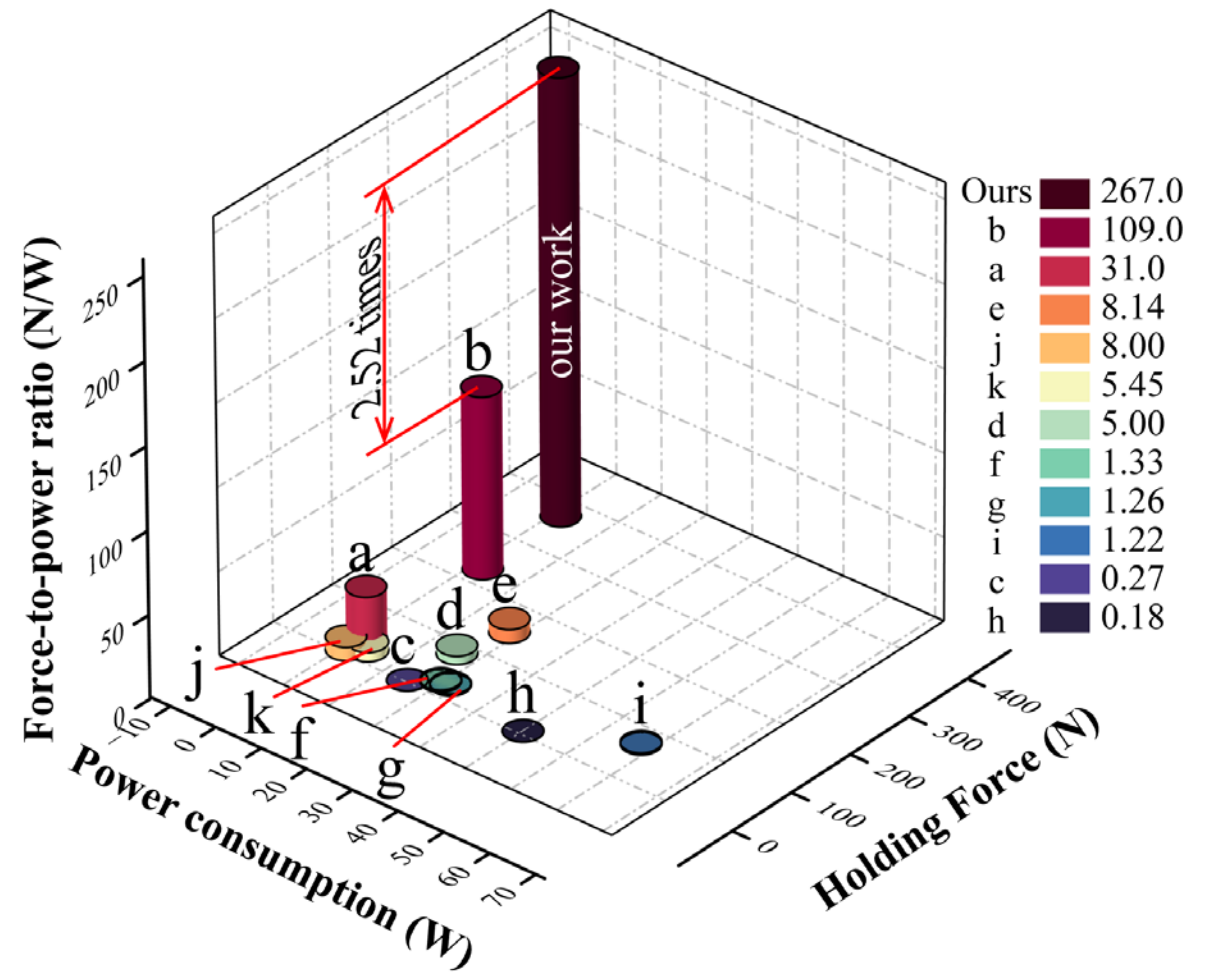

**Figure 10.** Comparative analysis between the MRGC with other reported actuators or clutches used in hand exoskeleton: Linear motor: a [53], d [54] and f [55]; SMA: g [56] and i [19]; Cable-driven: b [57], c [58] and e [59]; Pneumatic: h [60]; MR damper: j [13] and k [14].

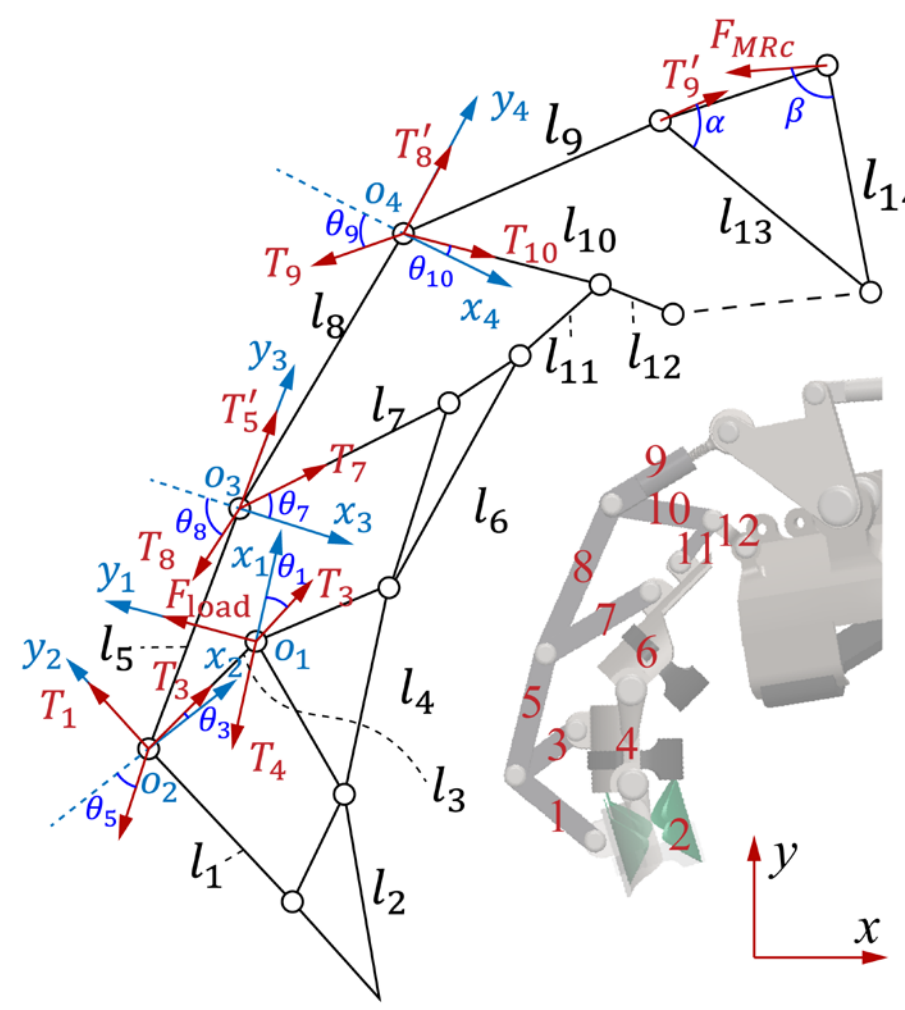

**Figure 11.** Kinetic model of the finger transmission mechanism.

$$T_3 = \frac{F_{load}}{sin\theta_1} \tag{7}$$

Similarly, in coordinate $x_2o_2y_2$, following equations can be yielded according to the equilibrium of forces:

$$T_5 \times sin\theta_5 + T_3 \times sin\theta_3 = T_1 \tag{8}$$

$$T_5 \times cos\theta_5 = T_3 \times cos\theta_3 \tag{9}$$

By substituting equations (7) and (8) into equation (9), we have

$$T_5 = \frac{F_{load} \times cos\theta_3}{sin\theta_1 \times cos\theta_5} \tag{10}$$

Using the similar force equilibrium method in coordinates $x_3o_3y_3$ and $x_4o_4y_4$ respectively, the relationship between the fingertip force $F_{load}$ and the force $F_{MRC}$ can be derived as:

$$\xi = \frac{\cos\theta_3\cos\theta_7\cos\theta_{10}}{\sin\theta_1\cos\theta_5\sin(\theta_8 - \theta_7)\sin(\theta_9 - \theta_{10})} \tag{11}$$

$$F_{MRC} = \frac{sin\alpha \cdot l_{13}}{sin\beta \cdot l_{14}} \cdot \xi \cdot F_{load} \tag{12}$$

where $l_{13} = 35\ mm$ and $l_{14} = 30\ mm$ are the lengths of the two levers of the V-shaped connection. $\theta_1, \theta_2, ..., \theta_{10}, \alpha$, and $\beta$ defined in Figure 11 have been measured in the grip endurance test and displayed in Table 3

**Table 3.** Parameter Value of the Finger Transmission

| Parameters | Value | Parameters | Value |
|---|---|---|---|
| $\theta_1, \theta_5$ | 34.0° | $\alpha$ | 49.4° |
| $\theta_3$ | 0° | $\beta$ | 110.7° |
| $\theta_7$ | 25.2° | $\theta_8 - \theta_7$ | 50.0° |
| $\theta_{10}$ | 6.9° | $\theta_9 - \theta_{10}$ | 43.1° |

By substituting the above angles and lengths into equation (12), we have

$$\begin{aligned} F_{load} &= \frac{sin\beta \cdot l_{14}}{sin\alpha \cdot l_{13}} \cdot \frac{1}{\xi} \cdot F_{MRC} \\ &\approx 0.285 \times F_{MRC} \end{aligned} \tag{13}$$

An MRHE consists of four fingers that are equipped with MRGCs; therefore, its support force, $\hat{F}_{support}$, can be expressed as follows by combining equations (4) and (13):

$$\begin{aligned} \hat{F}_{support} &\approx 0.285 \times 4 \times \begin{pmatrix} 8.336 + 35.412v \\ +375.650v^2 - 253.826v^3 \\ +59.948v^4 - 4.588v^5 \end{pmatrix} \\ &\approx 9.503 + 40.370v + 424.821v^2 \\ &- 289.362v^3 + 68.341v^4 \\ &- 5.197v^5 \end{aligned} \tag{14}$$

It can be calculated from the above equation that, when energized by the maximum recommended voltage of 2.0 V, the MRHE can provide approximately 419.79 N support force. When the MRHE is unpowered, given the initial damping force of MRGC is 7 N, the force required for one finger to drive the MRGC is calculated as 1.995 N using equation (13), which is far below a human finger can produce (around 10-40 N[61]). This demonstrates that the MRHE exhibits low resistance in passive mode, ensuring good drivability. Combined with its ability to deliver substantial support when activated, the proposed MRHE offers both ease of use and effective assistance.

### *3.3 Real-time power consumption monitoring of MRHE*

An experiment, as shown in Figure 12, was conducted to evaluate the total power consumption of the MRHE including four MRGCs and one control board. Each MRGC was supplied with a constant voltage of 2.0 V from the control board for 60 minutes. Power consumption data were recorded using a programmable DC power supply (DP100, ALIENTEK, China). Throughout the test, the average power of MRHE was found to be 5.65 W, with little power variations.

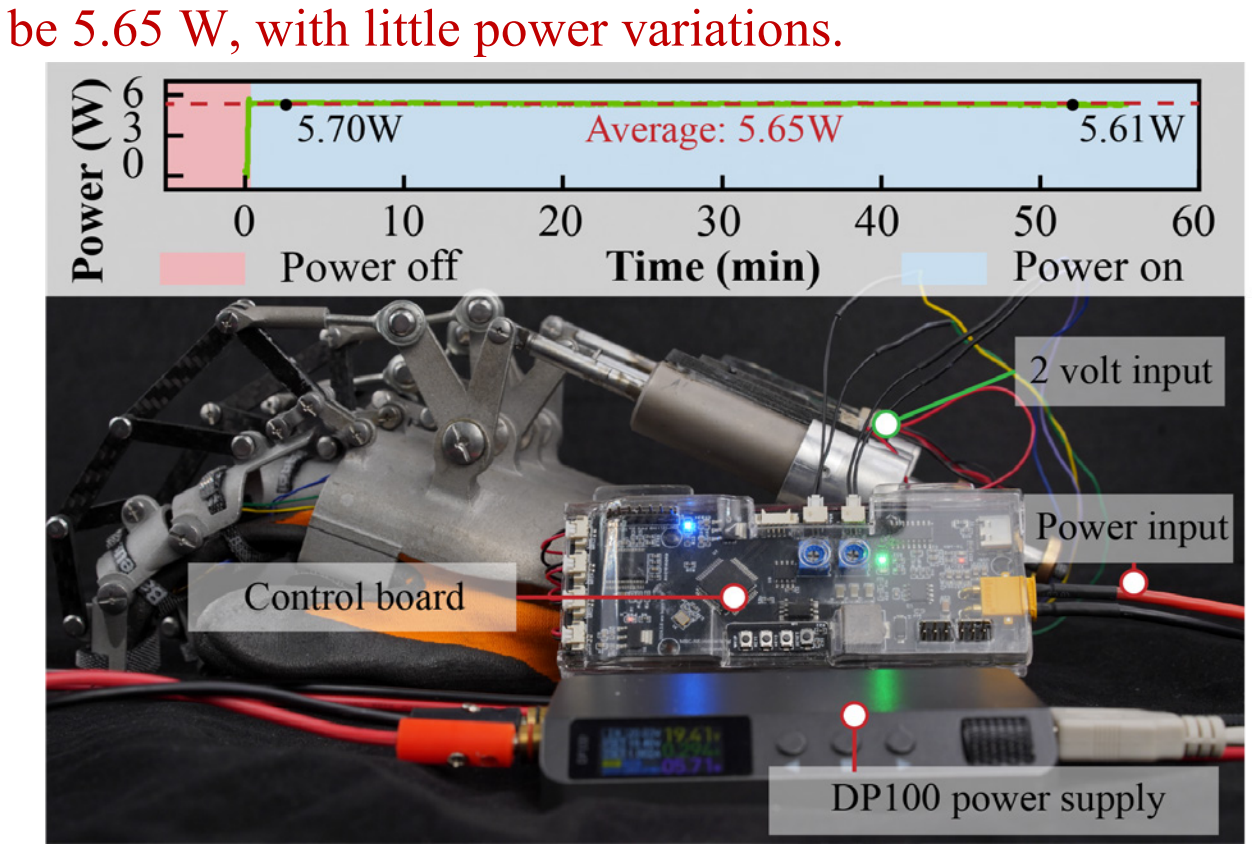

**Figure 12.** Experiment on power consumption monitoring of the MRHE.

## 4. Individual-wearing experiments

This section presents the individual-wearing experiments designed to assess the performance of MRHE in improving grip endurance. Three experiments were conducted, including a static gripping experiment, a dynamic weight-carrying experiment, and a repetitive weight-lifting experiment. The first experiment was conducted in an ideal and static condition, and the last two experiments were conducted in practical scenarios. Before participation, all subjects were provided written informed consent, and the study was approved by the Ethics Committee of the University of Science and Technology of China.

### *4.1 Static weight-gripping experiment*

The purpose of this experiment is to assess the effectiveness of the MRHE in enhancing human gripping endurance by conducting an ideally static weight-gripping experiment while minimizing interference from other body movements. A total of five subjects (three males and two females, with a mean mass of 61.1 ± 3 kg and a mean height of 1.70 ± 0.03 m, all of whom were right-handed) participated in this experiment. During the experiment, hand grip strength (HGS) is selected as a key index to evaluate muscle fatigue, as previous research [62] has indicated that muscle fatigue can affect hand grip strength (HGS). Meanwhile, the surface electromyographic (EMG) signal was recorded from the forearm muscles of subjects to further analyze the muscles' activity and fatigue.

### *4.1.1 Experimental process*

As shown in Figure 13(a), the experimental procedure for evaluating the enhancement of static grip endurance contains five steps. In the first step, each subject measures their max hand grip strength (HGS) three times using a grip dynamometer (SENSUN EH102R, China), following the guidelines set by the American Society of Hand Therapists (ASHT) [63]. To minimize the influence of fatigue, trials are spaced at 60-second intervals, and the average of the three measurements is recorded as the non-fatigue HGS. In the second step, subjects perform the grip endurance test while wearing the MRHE. Subsequently, in the third step, the max HGS was measured and marked as post-exoskeleton HGS. After a three-day break, subjects complete the grip endurance test without wearing MRHE in the fourth step, followed by a final HGS measurement in the fifth step, and acquire the Post-manual HGS.

Prior to the static grip endurance tests in the second and fourth steps, the experimenter assists the subject in donning the hand exoskeleton and makes necessary adjustments to ensure a secure fit. Then, the experimenter lifts a 17.5 kg dumbbell and places it in the subject's dominant hand, ensuring that the fingers of the subject maintain contact with the dumbbell while allowing the hand muscles to remain relaxed. The subject is requested to stand with their feet shoulder-width apart, holding the heavy object with the palm facing the body and looking straight ahead. At the start of the test, upon the experimenter's command, the assistant releases the heavy object, and the subject activates the auxiliary function of the MRHE to grasp the dumbbell for 1 minute. Figure 13(d) and (e) show the test with the subject wearing the MRHE. During the gripping tests (steps 2 and 4), surface electromyographic (sEMG) signals are measured from the forearm muscles using three EMG electrodes (Trigon Lab, DELSYS, America), which are attached to the forearms of the subject using specialized adhesive stickers and further secured with medical bandages to prevent movement. As shown in Figure 13(b) and (c), they are positioned on the brachioradialis, palmaris longus, and extensor digitorum muscles, key muscles involved in grip force generation. Prior to the electrode placement, the skin is cleansed with alcohol to ensure sensor contact quality. The sEMG signals were wirelessly transmitted to a personal computer and recorded using the EMGworks Acquisition software.

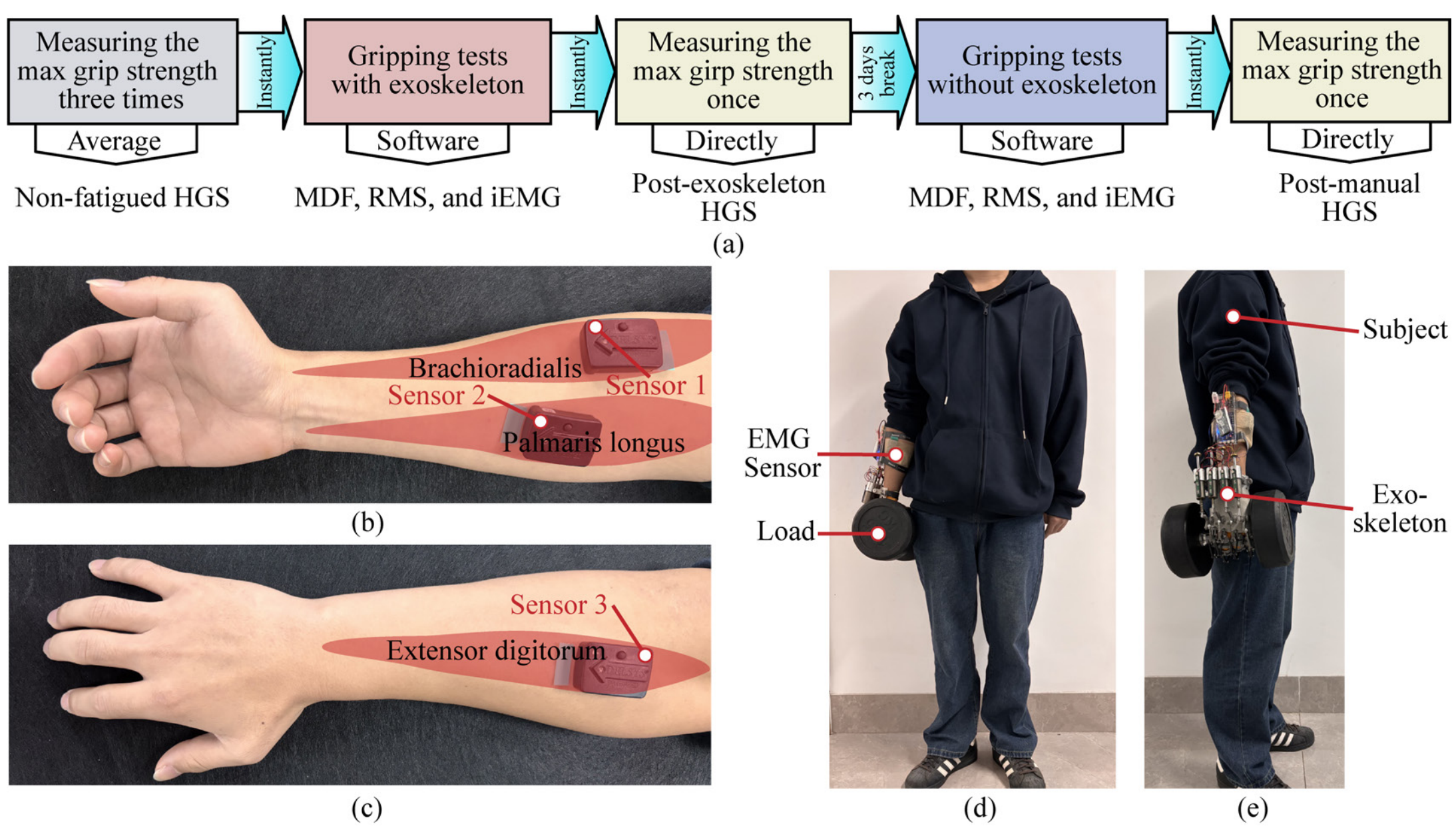

**Figure 13**. Setup for grip endurance experiments. (a) The procedure of the test. (b) EMG sensor placement of the palmar-side view. (c) EMG sensor placement of the dorsal-side view. (d) Frontal view during one grip endurance test. (e) Lateral view during one grip endurance test.

The data underwent offline processing using EMGworks Analysis software for visualization. The initial phase involved segmentation to isolate the relevant segments of interest. Subsequently, the data were filtered using a bandpass filter with a frequency range of 100 to 400 Hz to reduce noise interference. Linear envelope and integration techniques were then applied to extract the EMG amplitude, enabling the calculation of the root mean square (RMS) values and integrated electromyography (iEMG), respectively. RMS represents the average level of muscle fatigue, while iEMG reflects the cumulative muscle discharge and fatigue experienced over a specific period; both serve critical indexes in the time domain. In the frequency domain, the median frequency (MDF) acquired by frequency analysis is selected as a key indicator of muscle fatigue, with a low value denoting increased muscle fatigue [64].

### *4.1.2 Results*

The experimental results of the grip test are presented in Figure 14, where Figure 14(a-1) to (a-3) are the results of a male (Subject 1) and Figure 14(b-1) to (b-3) are the results of a female (Subject 2), in terms of MDF, RMS, and iEMG. The results of Subject 3-5 are provided in the Supplementary Material. Blue signals and bars are the results of the test without MRHE, while red signals and bars are the results acquired when the subjects are assisted by MRHE. It is seen from Figure 14(a-1) and (b-1) that the overall values of the MDF in the non-MRHE group are lower than in the with-MRHE group, especially in sensor 2 (palmaris longus), demonstrating that the non-MRHE group suffered increased muscle fatigue. As shown in Figures 14(a-2) and (b-2), the RMS signal for all sensors is obvious when subjects implement gripping motion without MRHE, indicating that muscles exert forces. In contrast, the RMS signal is significantly mitigated with smaller values in the with-MRHE group, which demonstrates very low muscle contraction rates when the MRHE provides substantial supporting forces, reducing the activity of the muscle involved in gripping.

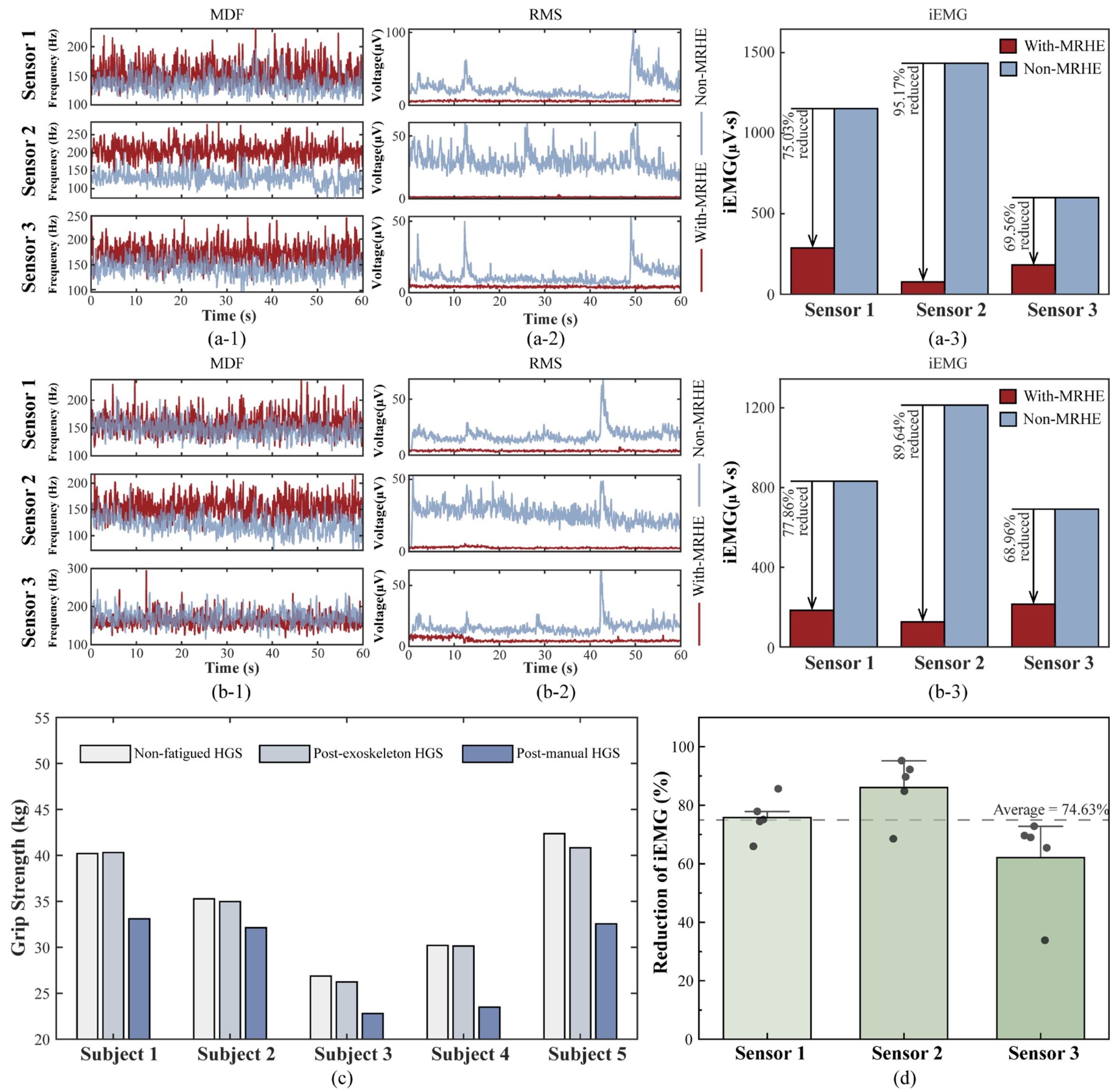

**Figure 14.** Experimental results of static weight gripping experiments. (a) One of the Male subjects (subject 1). (b) One of the Female subjects (subject 2). (c) Comparison of maximal grip strength under three conditions. (d) The average iEMG reduction of five subjects in terms of the three muscles after using the MRHE.

Over the period, the iEMG (Figure 14(a-3) and (b-3)) also shows a substantial decrease from the non-MRHE group to the with-MRHE group, ranging from 69.56% to 95.17% for males and 68.96% to 89.64% for females, while the reduction of sensor 2 (palmaris longus) presents the most apparent trend. As shown in Figure 14(d), the average reductions of iEMG values across five subjects are 75.75% for the brachioradialis (sensor 1), 86.04% for the palmaris longus (sensor 2), and 62.10% for the extensor digitorum (sensor 3), yielding an overall average reduction of 74.63%. These huge reductions fully verify the effectiveness of the MRHE in reducing the accumulated fatigue during static gripping tasks.

Meanwhile, the result of grip strength in Figure 14(c) also demonstrates the benefit of the MRHE. For all subjects, the post-manual HGS was reduced by an average of 17.41% compared to the non-fatigued HGS, indicating fatigue-induced strength loss. However, with MRHE support, the post-exoskeleton HGS only had a minor reduction of 0.71%, suggesting that the maximum grip strength of subjects was only slightly affected because the MRHE avoided muscle fatigue. Overall, similar results can be found in the two genders, and these findings underscore the exoskeleton's potential to effectively mitigate muscle fatigue and prevent injuries during extended or intense physical activities.

### *4.2 Practical applications in working*

This section presents the performance of the MRHE in dynamic scenarios, including carrying and lifting objects with interference like body movement and body posture.

#### *4.2.1 Dynamic weight-carrying experiment*

As shown in Figure 15, the experiment requires a subject (age 23, male, height 174.5 cm, weight 68.4 kg) to carry an aluminum extrusion (9 kg, 1000 mm×100 mm×100 mm) while walking on a treadmill at a speed of 3 km/h, simulating practical carrying scenarios of workers. The subject used his dominant hand (right hand) to hold the aluminum extrusion at its center of mass, while the non-dominant hand interacted with the aluminum extrusion using only two fingers to minimize its influence. The EMG electrodes are placed in the same way as previous static grip endurance tests. At the start of the experiment, the subject initiates treadmill walking. Once the designated speed (3m/s) is reached, the experimenter hands the aluminum extrusion to the subject and simultaneously starts recording the EMG signals. The walking and carrying task lasts for 30 seconds, after which the subject sets down the aluminum extrusion and ceases movement.

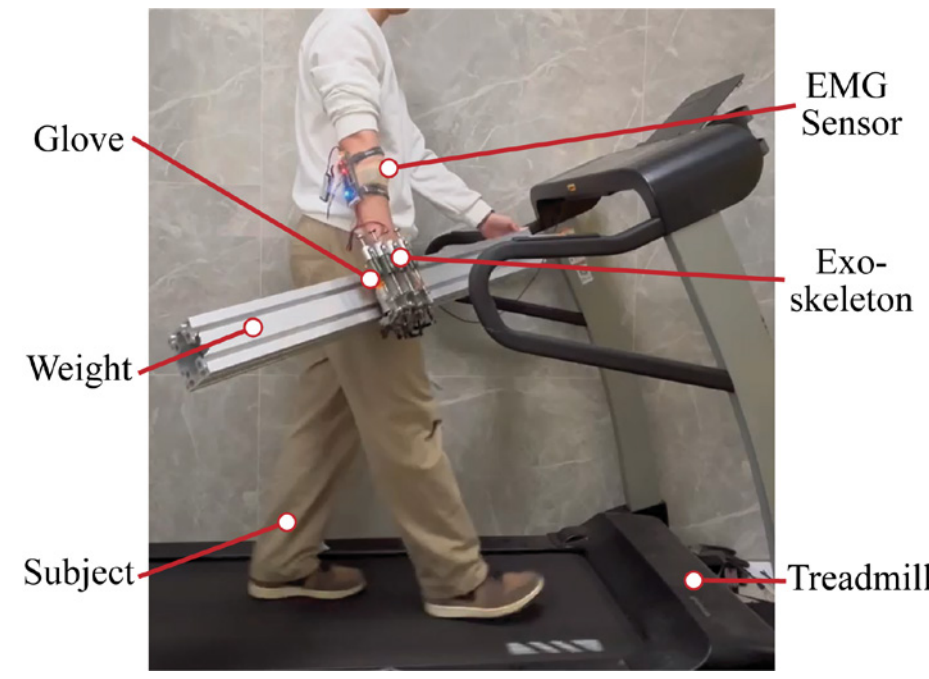

**Figure 15.** Experimental setup for the dynamic weight-carrying test.

Experiments with the assistance of MRHE and without are compared in Figure 16. As shown in Figure 16(a), the MDF value of the Brachioradialis muscle (Sensor 2) decreased dramatically in the absence of the exoskeleton, indicating an increase in muscle fatigue. Figure 16(b) shows that the RMS value was notably lower when the exoskeleton was used compared to when it was not. In Figure 16(c), the iEMG values for the three muscles decreased significantly by 70.42%, 88.66%, and 75.50%, respectively, when using MRHE, suggesting a substantial reduction in cumulative muscle fatigue. These results indicate that the proposed MRHE is highly effective in enhancing muscle endurance during dynamic activities even under the interference of movement.

#### *4.2.2 Repetitive weight-lifting experiment*

To assess the impact of the exoskeleton on the grip ability and endurance of the hand under sustained load, an experiment simulating a repetitive lifting task in the factory environment was conducted, as illustrated in Figure 17.

Prior to the experiment, a male subject (aged 22, 175.6 cm in height, and 72 kg in weight) performed preparatory exercises under the supervision of the experimenter. Initially, the subject positioned their feet shoulder-width apart, with the experimenter pre-marking the floor to ensure consistent stance alignment relative to the load across all trials. The subject then bent their knees and used their dominant hand to lift one end of an aluminum extrusion (20 kg, 2500 mm×100 mm×100 mm).

EMG sensors were applied in the same manner as in previous experiments. After the preparatory phase, the subject lifted the extrusion eight times without intervals upon the experimenter's command. During each repetition, the subject lifted the aluminum extrusion to its maximum height before lowering it, maintaining a steady lifting speed regardless of the presence of the exoskeleton. Consistency in lifting speed was emphasized for each repetition. Experiments were conducted when the subject was wearing and without wearing MRHE were conducted, and the data from EMG sensors were collected. In contrast to the above two experiments, the accumulation of muscle fatigue during repeated lifting was not significant enough to notably affect the alterations in MDF. Consequently, this study focused primarily on both qualitative and quantitative analyses of the RMS and iEMG values presented in Figure 18.

As shown in Figure 18(a), the application of the exoskeleton led to a significant reduction in RMS values, indicating a decrease in the intensity of muscle activity. Moreover, Figure 18(b) shows that the iEMG values of the three muscles corresponding to Sensor1, Sensor2, and Sensor3 decreased by 51.54%, 37.49%, and 42.23%, respectively, which are lower than the other two experiments due to lighter weight and the horizontal component of force. However, it still proves the exoskeleton's efficacy in reducing muscle fatigue and improving muscle endurance, especially in scenarios involving continuous load and repetitive strain.

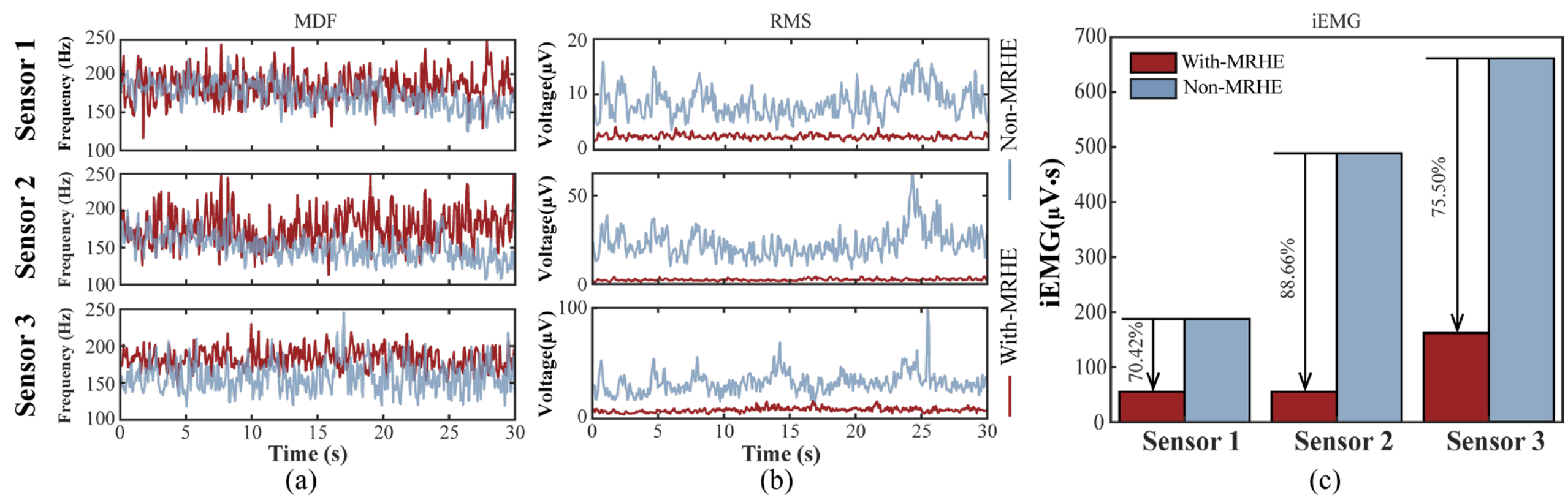

**Figure 16.** Experimental results of the dynamic weight-carrying experiment. (a) MDF values. (b) RMS values. (c) iEMG data.

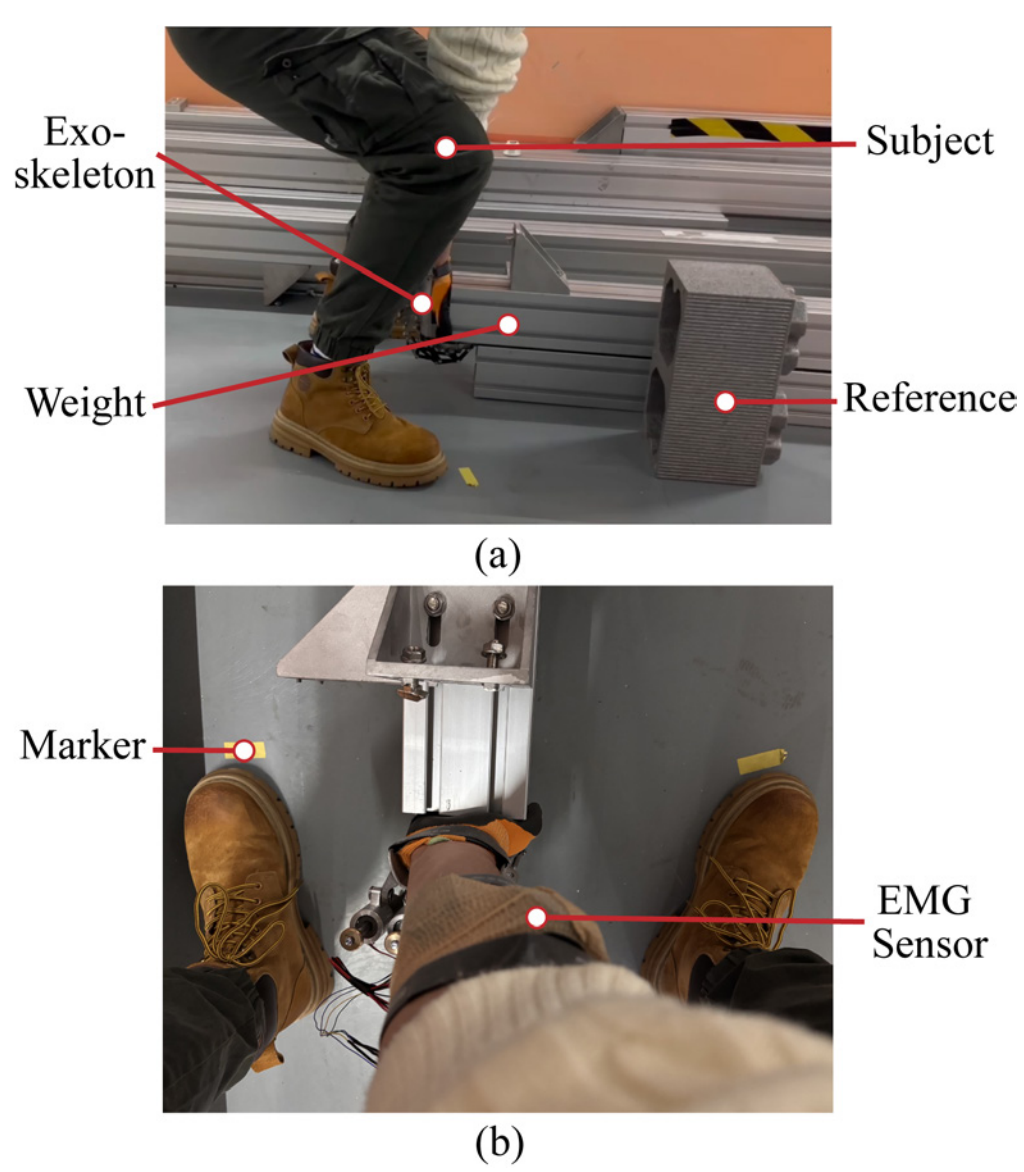

**Figure 17.** Experimental setup for the repetitive weight-lifting experiment. (a) Side view. (b) Top view.

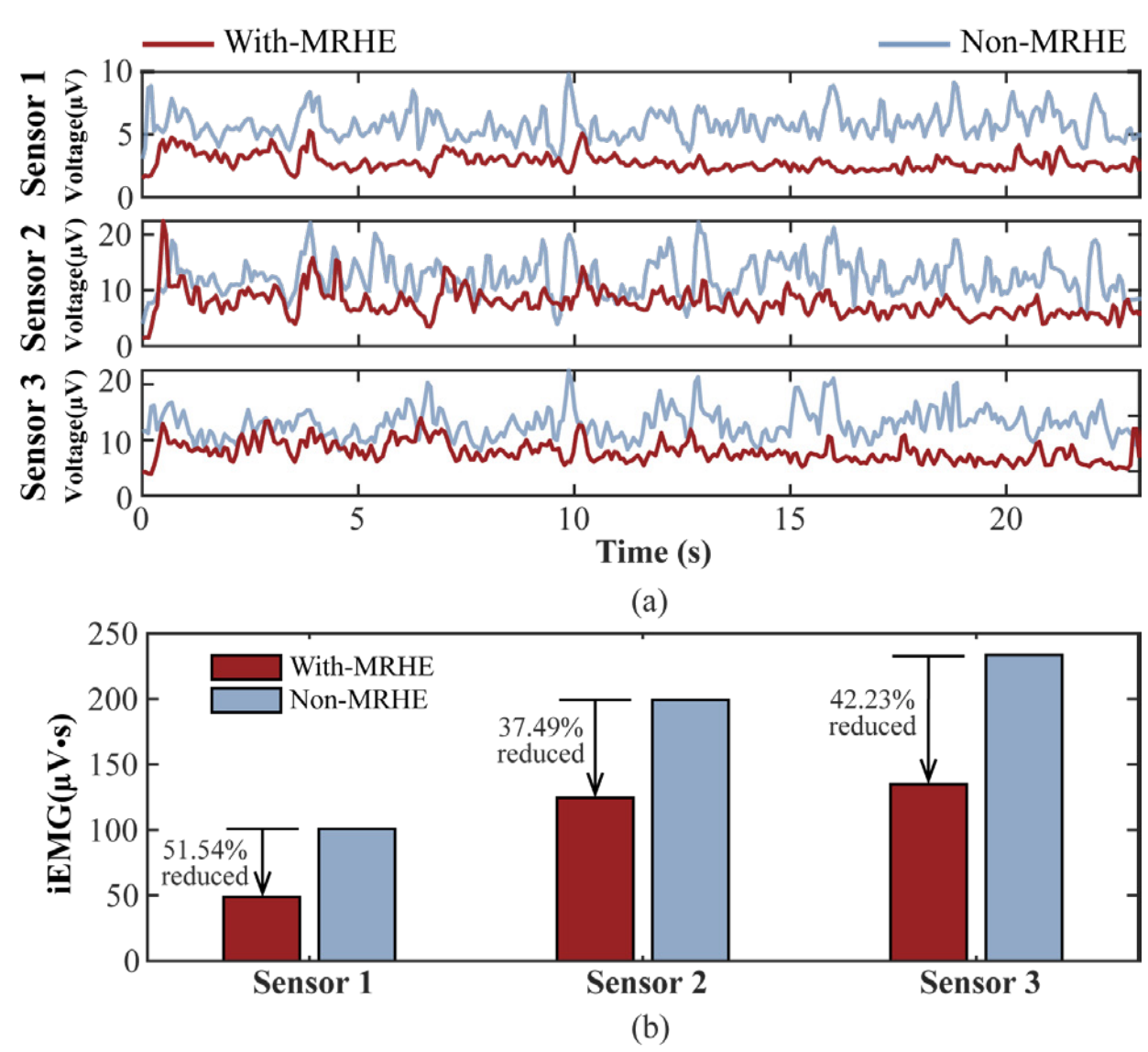

**Figure 18.** Experimental results of the repetitive weight-lifting experiment. (a) RMS values. (b) iEMG data.

## 5. Conclusion

This study presents a MRHE featuring high integration and a superior force-to-power ratio, designed to enhance grip endurance. The core innovation lies in the magnetorheological grease clutch (MRGC), which integrates a micro roller-enhanced structure to achieve high holding force with low power consumption. Experimental results show that the MRGC can deliver 381.15 N at 1.38 W, yielding a force-to-power ratio of 276.18 N/W, which is 2.35 times higher than that of previous reported actuators. This allows the MRHE to provide a support force of 419.79 N under 2.0 V input. Meanwhile, the compact design of MRGC substantially improves the MRHE's portability and level of integration.

Human trials validated the MRHE's effectiveness in fatigue mitigation in both static and dynamic experiments. The static weight-gripping experiment showed that the cumulative muscle discharge and fatigue, denoted by the iEMG value, were reduced by 74.63% using the MRHE. In dynamic tests, including weight-carrying and weight-lifting experiments, iEMG values were mitigated by at least 70.42% and 37.49%, respectively. These results indicate the potential application of the MRHE in practical scenarios.

## Acknowledgments

This work was supported by National Natural Science Foundation of China (Grant No. U21A20119 and 52105081), Major Project of Anhui Province's Science and Technology Innovation Breakthrough Plan (202423h08050003), Anhui's Key R&D Program of China (Grant No. 202104a05020009), USTC start-up funding (Grant No. KY2090000067), and the Fundamental Research Funds for the Central Universities (SC5290005162).

## Data availability statement

All data that support the findings of this study are included within the article (and any supplementary files).

## Ethical statement

The human research participants consented to all the experiments and identifiable images in this work. Experimental exercises on the individuals were conducted by protocols approved by the Ethical Committee of The First Affiliated Hospital of the University of Science and Technology of China (Number 2023KY239).

## References

[1] Rogers W A, Mitzner T L and Bixter M T 2020 Understanding the potential of technology to support enhanced activities of daily living (EADLs) *Gerontechnology* **19** 125–37

[2] White C, Dixon K, Samuel D and Stokes M 2013 Handgrip and quadriceps muscle endurance testing in young adults *SpringerPlus* **2** 451

[3] Nicolay C W and Walker A L 2005 Grip strength and endurance: Influences of anthropometric variation, hand dominance, and gender *International Journal of Industrial Ergonomics* **35** 605–18

[4] Krishnan K S, Raju G and Shawkataly O 2021 Prevalence of Work-Related Musculoskeletal Disorders: Psychological and Physical Risk Factors *International Journal of Environmental Research and Public Health* **18** 9361

[5] Wu N and Xie S Q 2024 Adaptation of hand exoskeletons for occupational augmentation: A literature review *Robotics and Autonomous Systems* **174** 104618

[6] Dupont P E, Nelson B J, Goldfarb M, Hannaford B, Menciassi A, O'Malley M K, Simaan N, Valdastri P and Yang G-Z 2021 A decade retrospective of medical robotics research from 2010 to 2020 *Science Robotics* **6** eabi8017

[7] Heo P, Gu G M, Lee S, Rhee K and Kim J 2012 Current hand exoskeleton technologies for rehabilitation and assistive engineering *Int. J. Precis. Eng. Manuf.* **13** 807–24

[8] Shahid T, Gouwanda D, Nurzaman S G and Gopalai A A 2018 Moving toward Soft Robotics: A Decade Review of the Design of Hand Exoskeletons *Biomimetics* **3** 17

[9] Soekadar S R, Witkowski M, Gómez C, Opisso E, Medina J, Cortese M, Cempini M, Carrozza M C, Cohen L G, Birbaumer N and Vitiello N 2016 Hybrid EEG/EOG-based brain/neural hand exoskeleton restores fully independent daily living activities after quadriplegia *Science Robotics* **1** eaag3296

[10] Li H and Cheng L 2024 A systematic review on hand exoskeletons from the mechatronics aspect *IEEE/ASME Trans. Mechatronics* 1–19

[11] Du Plessis T, Djouani K and Oosthuizen C 2021 A Review of Active Hand Exoskeletons for Rehabilitation and Assistance *Robotics* **10** 40

[12] Kim S J, Kim Y, Lee H, Ghasemlou P and Kim J 2018 Development of an MR-compatible hand exoskeleton that is capable of providing interactive robotic rehabilitation during fMRI imaging *Med. Biol. Eng. Comput.* **56** 261–72

[13] Yi A, Zahedi A, Wang Y, Tan U-X and Zhang D 2019 A novel exoskeleton system based on magnetorheological fluid for tremor suppression of wrist joints *2019 IEEE 16th International Conference on Rehabilitation Robotics (ICORR)* 2019 IEEE 16th International Conference on Rehabilitation Robotics (ICORR) pp 1115–20

[14] Cassar D J and Saliba M A A force feedback glove based on magnetorheological fluid: Prototype development and evaluation
[15] Sarac M, Solazzi M and Frisoli A 2019 Design Requirements of Generic Hand Exoskeletons and Survey of Hand Exoskeletons for Rehabilitation, Assistive, or Haptic Use *IEEE Transactions on Haptics* **12** 400–13
[16] Ben-Tzvi P and Ma Z 2015 Sensing and Force-Feedback Exoskeleton (SAFE) Robotic Glove *IEEE Transactions on Neural Systems and Rehabilitation Engineering* **23** 992–1002
[17] Hong M B, Kim S J, Ihn Y S, Jeong G-C and Kim K 2019 KULEX-Hand: An Underactuated Wearable Hand for Grasping Power Assistance *IEEE Trans. Robot.* **35** 420–32
[18] Anon A Modular Wearable Finger Interface for Cutaneous and Kinesthetic Interaction: Control and Evaluation | IEEE Journals & Magazine | IEEE Xplore
[19] Villoslada Á, Rivera C, Escudero N, Martín F, Blanco D and Moreno L 2019 Hand Exo-Muscular System for Assisting Astronauts During Extravehicular Activities *Soft Robotics* **6** 21–37
[20] Ho N S K, Tong K Y, Hu X L, Fung K L, Wei X J, Rong W and Susanto E A 2011 An EMG-driven exoskeleton hand robotic training device on chronic stroke subjects: Task training system for stroke rehabilitation *2011 IEEE International Conference on Rehabilitation Robotics* 2011 IEEE International Conference on Rehabilitation Robotics pp 1–5
[21] Jo I and Bae J 2017 Design and control of a wearable and force-controllable hand exoskeleton system *Mechatronics* **41** 90–101
[22] Dragusanu M, Iqbal M Z, Baldi T L, Prattichizzo D and Malvezzi M 2022 Design, Development, and Control of a Hand/Wrist Exoskeleton for Rehabilitation and Training *IEEE Transactions on Robotics* **38** 1472–88
[23] Anon Fully Wearable Actuated Soft Exoskeleton for Grasping Assistance in Everyday Activities | Soft Robotics
[24] Cempini M, Cortese M and Vitiello N 2015 A Powered Finger–Thumb Wearable Hand Exoskeleton With Self-Aligning Joint Axes *IEEE/ASME Transactions on Mechatronics* **20** 705–16
[25] Laffranchi M, Boccardo N, Traverso S, Lombardi L, Canepa M, Lince A, Semprini M, Saglia J A, Naceri A, Sacchetti R, Gruppioni E and De Michieli L 2020 The Hannes hand prosthesis replicates the key biological properties of the human hand *Sci. Robot.* **5** eabb0467
[26] Gasser B W, Bennett D A, Durrough C M and Goldfarb M 2017 Design and preliminary assessment of Vanderbilt hand exoskeleton *2017 International Conference on Rehabilitation Robotics (ICORR)* 2017 International Conference on Rehabilitation Robotics (ICORR) pp 1537–42
[27] Bruder D, Graule M A, Teeple C B and Wood R J 2023 Increasing the payload capacity of soft robot arms by localized stiffening *Science Robotics* **8** eadf9001
[28] Morales R, Badesa F J, García-Aracil N, Sabater J M and Pérez-Vidal C 2011 Pneumatic robotic systems for upper limb rehabilitation *Med Biol Eng Comput* **49** 1145–56
[29] Wehner M, Tolley M T, Mengüç Y, Park Y-L, Mozeika A, Ding Y, Onal C, Shepherd R F, Whitesides G M and Wood R J 2014 Pneumatic Energy Sources for Autonomous and Wearable Soft Robotics *Soft Robotics* **1** 263–74
[30] Sreekumar M, Nagarajan T, Singaperumal M, Zoppi M and Molfino R 2007 Critical review of current trends in shape memory alloy actuators for intelligent robots ed Derby S *Ind. Robot: Int. J.* **34** 285–94
[31] Anon Actuation Technologies for Soft Robot Grippers and Manipulators: A Review | Current Robotics Reports
[32] Carey A J and Robinson S 2016 An Unpowered Exoskeleton to Reduce Astronaut Hand Fatigue during Microgravity EVA *AIAA SPACE 2016* AIAA SPACE 2016 (Long Beach, California: American Institute of Aeronautics and Astronautics)
[33] Refour E M, Sebastian B, Chauhan R J and Ben-Tzvi P 2019 A General Purpose Robotic Hand Exoskeleton With Series Elastic Actuation *Journal of Mechanisms and Robotics* **11** 060902
[34] Anon A force augmenting exoskeleton for the human hand designed for pinching and grasping | IEEE Conference Publication | IEEE Xplore
[35] Yang J, Sun S, Yang X, Ma Y, Yun G, Chang R, Tang S-Y, Nakano M, Li Z and Du H 2022 Equipping new SMA artificial muscles with controllable MRF exoskeletons for robotic manipulators and grippers *IEEE/ASME Transactions on Mechatronics* **27** 4585–96
[36] Dai J, Chang H, Zhao R, Huang J, Li K and Xie S 2019 Investigation of the relationship among the microstructure, rheological properties of MR grease and the speed reduction performance of a rotary micro-brake *Mechanical Systems and Signal Processing* **116** 741–50
[37] Yi A, Zahedi A, Wang Y, Tan U-X and Zhang D 2019 A Novel Exoskeleton System Based on Magnetorheological Fluid for Tremor Suppression of Wrist Joints *2019 IEEE 16th International Conference on Rehabilitation Robotics (ICORR)* 2019 IEEE 16th International Conference on Rehabilitation Robotics (ICORR) (Toronto, ON, Canada: IEEE) pp 1115–20
[38] Mousavi S H and Sayyaadi H 2018 Optimization and testing of a new prototype hybrid MR brake with arc form surface as a prosthetic knee *IEEE/ASME Trans. Mechatronics* **23** 1204–14
[39] Nam T H and Ahn K K 2009 New approach to designing an MR brake using a small steel roller and MR fluid *J Mech Sci Technol* **23** 1911–23
[40] Mai X, Yang J, Li L, Zi B, Zhang S, Gong X, Li W, Yun G and Sun S 2025 Non-motorized hand exoskeleton for rescue and beyond: Substantially elevating grip endurance and strength *IEEE Trans. Robot.* 1–15
[41] Ye X, Dong J, Wu B, Qing O, Wang J and Zhang G 2024 Design and modeling of a double rod magnetorheological grease damper *J Mech Sci Technol* **38** 4065–75
[42] Sugiyama S, Sakurai T and Morishita S 2013 Vibration control of a structure using Magneto-Rheological grease damper *Front. Mech. Eng.* **8** 261–7
[43] Changsheng Z 2006 Experimental Investigation on the Dynamic Behavior of a Disk-type Damper based on Magnetorheological Grease *Journal of Intelligent Material Systems and Structures* **17** 793–9

[44] Eshgarf H, Ahmadi Nadooshan A and Raisi A 2022 An overview on properties and applications of magnetorheological fluids: Dampers, batteries, valves and brakes *Journal of Energy Storage* **50** 104648
[45] Fu J, Bai J, Lai J, Li P, Yu M and Lam H-K 2019 Adaptive fuzzy control of a magnetorheological elastomer vibration isolation system with time-varying sinusoidal excitations *Journal of Sound and Vibration* **456** 386–406
[46] Ding L, Pei L, Xuan S, Fan X, Cao X, Wang Y and Gong X 2020 Ultrasensitive Multifunctional Magnetoresistive Strain Sensor Based on Hair-Like Magnetization-Induced Pillar Forests *Advanced Electronic Materials* **6** 1900653
[47] Blake J and Gurocak H B 2009 Haptic glove with MR brakes for virtual reality *IEEE/ASME Trans. Mechatronics* **14** 606–15
[48] Anon 2021 Characteristic analysis and squeezing force mathematical model for magnetorheological fluid in squeeze mode *Journal of Magnetism and Magnetic Materials* **529** 167736
[49] Tang X, Zhang X, Tao R and Rong Y 2000 Structure-enhanced yield stress of magnetorheological fluids *J. Appl. Phys.* **87** 2634–8
[50] Zhang X Z, Gong X L, Zhang P Q and Wang Q M 2004 Study on the mechanism of the squeeze-strengthen effect in magnetorheological fluids *J. Appl. Phys.* **96** 2359–64
[51] Pei P and Peng Y 2021 The squeeze strengthening effect on the rheological and microstructured behaviors of magnetorheological fluids: a molecular dynamics study *Soft Matter* **17** 184–200
[52] Fowell M, Olver A V, Gosman A D, Spikes H A and Pegg I 2006 Entrainment and inlet suction: two mechanisms of hydrodynamic lubrication in textured bearings *J. Tribol.* **129** 336–47
[53] Nycz C J, Bützer T, Lambercy O, Arata J, Fischer G S and Gassert R 2016 Design and Characterization of a Lightweight and Fully Portable Remote Actuation System for Use With a Hand Exoskeleton *IEEE Robotics and Automation Letters* **1** 976–83
[54] Chen W, Li G, Li N, Wang W, Yu P, Wang R, Xue X, Zhao X and Liu L 2022 Soft Exoskeleton With Fully Actuated Thumb Movements for Grasping Assistance *IEEE TRANSACTIONS ON ROBOTICS* **38**
[55] Anon High force density linear permanent magnet motors : “electromagnetic muscle actuators”
[56] Tang T, Zhang D, Xie T and Zhu X 2013 An exoskeleton system for hand rehabilitation driven by shape memory alloy *2013 IEEE International Conference on Robotics and Biomimetics (ROBIO)* 2013 IEEE International Conference on Robotics and Biomimetics (ROBIO) pp 756–61
[57] Dittli J, Hofmann U A T, Bützer T, Smit G, Lambercy O and Gassert R 2021 Remote Actuation Systems for Fully Wearable Assistive Devices: Requirements, Selection, and Optimization for Out-of-the-Lab Application of a Hand Exoskeleton *Front. Robot. AI* **7** 596185
[58] Marconi D, Baldoni A, McKinney Z, Cempini M, Crea S and Vitiello N 2019 A novel hand exoskeleton with series elastic actuation for modulated torque transfer *Mechatronics* **61** 69–82
[59] Hofmann U A T, Bützer T, Lambercy O and Gassert R 2018 Design and Evaluation of a Bowden-Cable-Based Remote Actuation System for Wearable Robotics *IEEE Robotics and Automation Letters* **3** 2101–8
[60] Ramos O, Múnera M, Moazen M, Wurdemann H and Cifuentes C A 2022 Assessment of Soft Actuators for Hand Exoskeletons: Pleated Textile Actuators and Fiber-Reinforced Silicone Actuators *Front. Bioeng. Biotechnol.* **10** 924888
[61] Li Z-M, Pfaeffle H J, Sotereanos D G, Goitz R J and Woo S L-Y 2003 Multi-directional strength and force envelope of the index finger *Clin. Biomech.* **18** 908–15
[62] Quattrocchi A, Garufi G, Gugliandolo G, De Marchis C, Collufio D, Cardali S M and Donato N 2024 Handgrip Strength in Health Applications: A Review of the Measurement Methodologies and Influencing Factors *Sensors* **24** 5100
[63] Innes E 1999 Handgrip strength testing: A review of the literature *Aus Occup Therapy J* **46** 120–40
[64] Fernando J B, Yoshioka M and Ozawa J 2016 Estimation of muscle fatigue by ratio of mean frequency to average rectified value from surface electromyography *2016 38th Annual International Conference of the IEEE Engineering in Medicine and Biology Society (EMBC)* 2016 38th Annual International Conference of the IEEE Engineering in Medicine and Biology Society (EMBC) (Orlando, FL, USA: IEEE) pp 5303–6